\documentclass[letterpaper]{article} 
\usepackage{aaai23}
\usepackage{times}  
\usepackage{helvet}  
\usepackage{courier}  
\usepackage[hyphens]{url}  
\usepackage{graphicx} 
\usepackage{natbib}  
\usepackage{caption} 
\usepackage{algorithm}

\usepackage{setspace}
\usepackage{amsmath}
\usepackage{amssymb}
\DeclareUnicodeCharacter{2212}{−}
\usepackage{import}

\usepackage{algpseudocode}

\usepackage{longtable}
\usepackage{multirow}
\usepackage{booktabs}
\usepackage[acronym]{glossaries}
\usepackage[htt]{hyphenat}
\usepackage{caption}
\usepackage{subcaption}
\usepackage{import}

\usepackage{newfloat}
\usepackage{listings}
\DeclareCaptionStyle{ruled}{labelfont=normalfont,labelsep=colon,strut=off} 
\floatstyle{ruled}
\newfloat{listing}{tb}{lst}{}
\floatname{listing}{Listing}
\title{Utilizing Semantic Textual Similarity for Clinical Survey Data Feature Selection} 
\author{
    Benjamin C. Warner\textsuperscript{\rm 1},
    Ziqi Xu\textsuperscript{\rm 1},
    Simon Haroutounian\textsuperscript{\rm 2},
    Thomas Kannampallil\textsuperscript{\rm 1,2},
    Chenyang Lu\textsuperscript{\rm 1}
}
\affiliations{
    \textsuperscript{\rm 1}Washington University in St. Louis\\

    \textsuperscript{\rm 2}Washington University School of Medicine in St. Louis\\
    \{b.c.warner, ziqixu, sharout, thomas.k, lu\}@wustl.edu
}

\newacronym{sts}{STS}{semantic textual similarity}
\newacronym{mi}{MI}{mutual information}
\newacronym{auroc}{AUROC}{area under the receiver-operator curve}
\newacronym{auprc}{AUPRC}{area under the precision-recall curve}
\newacronym{mrmr}{mRMR}{minimal-redundancy-maximal-relevance}
\newacronym{snn}{SNN}{siamese neural network}
\newacronym{cv}{CV}{cross-validation}
\newacronym{lm}{LM}{language model}
\newacronym{llm}{LLM}{large language model}
\newacronym{bert}{BERT}{Bidirectional Encoder Representations from Transformers}
\newacronym{redcap}{REDCap}{Research Electronic Data Capture}
\newacronym{svm}{SVM}{support vector machine}
\newacronym{ppsp}{PPSP}{persistent post-surgical pain}
\newacronym{gnb}{Gaussian NB}{Gaussian Na\"{i}ve Bayes}
\newacronym{ml}{ML}{machine learning}
\newacronym{mqo}{mRMR-s}{mRMR with semantic textual similarity}
\newacronym{mb}{mRMR-i}{mRMR with mutual information}
\newacronym{mc}{mRMR-h}{mRMR with hybrid mutual information and semantic textual similarity}
\newacronym{mlp}{MLP}{multilayer perceptron}
\newacronym{knn}{$k$-NN}{$k$-nearest neighbors}
\newacronym{nlp}{NLP}{natural language processing}
\newacronym{shap}{SHAP}{SHapley Additive exPlanations}

\begin{document}

\maketitle

\begin{abstract}
Survey data can contain a high number of features while having a comparatively low quantity of examples. \Acrlong{ml} models that attempt to predict outcomes from survey data under these conditions can overfit and result in poor generalizability. One remedy to this issue is feature selection, which attempts to select an optimal subset of features to learn upon. A relatively unexplored source of information in the feature selection process is the usage of textual names of features, which may be semantically indicative of which features are relevant to a target outcome. The relationships between feature names and target names can be evaluated using \glspl{lm} to produce \gls{sts} scores, which can then be used to select features. We examine the performance using \gls{sts} to select features directly and in the \acrfull{mrmr} algorithm. The performance of \acrshort{sts} as a feature selection metric is evaluated against preliminary survey data collected as a part of a clinical study on \acrfull{ppsp}. The results suggest that features selected with \gls{sts} can result in higher performance models compared to traditional feature selection algorithms.
\end{abstract}

\section{Introduction}

Because of the high cost of collecting surveys in human subjects research, survey data can suffer from high dimensionality and an relatively inadequate number of examples to learn from. One solution is to select a subset of features when training an \acrshort{ml} model, but because this relies upon the already small data to learn from, suboptimal selections can be made. This issue is particularly important for clinical data, where the true relationships between features and labels may be too complex or unknown to make optimal decisions regarding feature choices. We explore this problem through a collection of surveys collected to assess \gls{ppsp}, where a patient experiences surgically-related pain for longer than expected \cite{vila2020cognitive}, and whose exact causes are presently unclear \cite{haroutiunian2013neuropathic}. 

\subsection{Proposed Approach}

Survey answers are the result of questions that have text that may be semantically related to a target outcome, and in turn semantically similar or dissimilar to one another. Intuition suggests that \gls{sts} is a useful analogue to statistical measures that capture relationships between features, such as \gls{mi}. \Gls{sts} may then be useful for determining which questions are \textit{relevant} to predicting a target question. Moreover, \gls{sts} scores may be useful in determining which questions are \textit{redundant} to each other. This may be especially helpful for smaller datasets where there is a limited amount of information immediately available to learn from. To this end, we evaluate the use of \gls{sts} scores directly to determine the most relevant features, and test \gls{sts} scores both as a direct replacement and compliment in algorithms utilizing statistical scores, such as \gls{mrmr}.

There appears to be nearly no literature examining feature selection with embeddings or \gls{sts}. The closest match to our knowledge examined the usage of \texttt{word2vec} continuous bag-of-words embeddings \cite{mikolov2013efficient} trained upon Twitter data to select Google search query trends matching the embeddings of a target concept \cite{lampos2017enhancing}. Our analysis differs in several key ways, with the principal difference being that we evaluate selection using the top $N$ \gls{sts} scores and \gls{mrmr}, in addition to the procedure in \citet{lampos2017enhancing}, where they features select beyond a $k$-standard deviation threshold. Another major difference is the usage of \glspl{llm} to calculate scores, which is a more recent model than \texttt{word2vec}, and demonstrably perform better with regards to general and clinical tasks \cite{roy2021incorporating}. 
Finally, \citet{lampos2017enhancing} apply their filtration method to a dataset with 35,572 different time-series features, whereas we evaluate a smaller tabular dataset in both examples and dimensions.

We present several empirical contributions in this paper:
\begin{itemize}
    \item An examination of how \gls{sts} scores generated by \glspl{lm} between feature and target questions can help select features, either alone or in combination with statistical measures such as \gls{mi}.
    \item Evaluating the performance of \gls{sts} scores produced by different models in feature selection.
    \item A demonstration of how \gls{sts}-based feature selection algorithms can reduce overfitting.
\end{itemize}

We make our code available at \path{https://github.com/bcwarner/sts-select}, and  via \texttt{pip install sts-select}.

\section{Related Work}

\subsection{Feature Selection} \label{sec:intro_fs}

Fitting high-dimensional data is particularly difficult when the number of examples is low---as is with clinical data collected from human subjects---since a model can easily overfit on the training data. To counter this, we can employ the strategy of \textit{feature selection}, where a subset of the overall features in a dataset are selected for learning. 

Feature selection methods can be divided into three categories: \textit{embedded}, \textit{wrapper}, and \textit{filter} methods. Embedded methods incorporate feature selection as a part of training, while wrapper methods interact in a feedback loop with the learning model. Filter methods select a subset of features based on properties of the dataset before the model is able to learn on the dataset, which differs from embedded and wrapper methods in that they do not form a feedback loop with the model \cite{guyon2008feature}. Because of their independence, they tend to generalize well \cite{remeseiro2019review}.

Feature selection methods used in clinical survey data cover a broad range of techniques. A study examining autism spectrum disorder (ASD) survey data, examined feature selection using principal component analysis, t-distributed stochastic neighbor embedding, and denoising autoencoders; and also found that survey features targeting ASD tend to have high levels of redundancy \cite{washington2019feature}. Some of the other feature selection methods found for models involve questionnaires include wrapper models based on random forests \cite{niemann2020development}, bootstrapped feature selection \cite{abbas2018machine},
principal component analysis, multicluster feature selection \cite{saridewi2020feature}, permutation importance \cite{chen2023feature}, and ReliefF \cite{abut2016developing}. 

One particularly useful feature selection technique for is \acrfull{mrmr} , which aims to maximize the \textit{relevance} of features to the target, while minimizing the \textit{redundancy} between selected features. This is particularly useful when we have a small number of features that are correlated and want to ensure a model incorporates as broad as a set of information as possible \cite{peng2005feature, ramirez2017fast}. 

Underpinning the \gls{mrmr} objective function is the \acrfull{mi} between classes and features, which measures the amount of shared information between two distributions. 

Calculating true \gls{mi} between two features is computationally costly, but can be approximated using one of several methods. The \gls{mi} approximation methods used in this paper are from scikit-learn \cite{pedregosa2011scikit}, and are a synthesis of the $k$-nearest neighbors approaches described in \cite{kraskov2004estimating, ross2014mutual}.

\subsection{\Acrlong{sts}}

\Acrfull{sts} is a task where a \gls{lm} is used to score the semantic similarity of two sentences, generally by evaluating the differences between embeddings generated by a model. Cosine similarity is one typical function used to compute the distance between embeddings \cite{reimers2019sentence, oniani2022few}.

\gls{sts} scores can be produced with traditional \glspl{lm} such as \texttt{word2vec} \cite{mikolov2013efficient} or FastText \cite{joulin2016fasttext}. They may also be produced with \acrfullpl{llm}, which are a class of \gls{lm} that is broadly defined as having on the order of high millions or more of parameters; and have been typically built with the transformer architecture \cite{manning2022human}. 

\Glspl{llm} typically employ the \textit{pre-training}/\textit{fine-tuning} paradigm, where a model is trained with an unsupervised learning task, and then modified and trained to complete a supervised task \cite{devlin2018bert, radford2019language}. Pre-training is particularly useful since it results in better generalization \cite{erhan2010does}, and because it means that computationally expensive pre-trained models can be reused for different tasks \cite{wolf2019huggingface}. \Glspl{llm} are distinct from traditional \glspl{lm} in that they have demonstrated capabilities at many reasoning tasks involving semantic meaning \cite{singhal2022large, wei2023overview}, and are highly applicable since survey features may involve more complex word relationships.

Clinical language involves vocabulary and semantic meaning that is often not present in non-clinical texts, and various pre-trained architectures exist to fill this gap. To date, there are over 80 clinical \glspl{lm} available \cite{wornow2023shaky}, with a diverse set of architectural and training designs. Clinical \glspl{lm} can perform better than general \glspl{lm} on tasks specific to the clinical domain \cite{alsentzer2019publicly}, and can do so more efficiently than a general-purpose models \cite{lehman2023we}.  

\section{Methodology}

\subsection{Scoring \& Selection}

A distinction can be made between the \textit{scoring} of features, where we measure a feature's relationship to a target or another feature, and the \textit{selection} of features, where we then use these scores to select the appropriate features. We evaluate the performance of two principal scoring tools, namely \gls{mi} and \gls{sts}. In addition, we evaluate the linear combination of them, with the coefficient for \gls{sts} being a hyperparameter $\alpha$, which we test over a logarithmically-spaced range of 30 values from $[10^{-2}, 10^{2}]$. For selection methods that employ these scores, we evaluate selecting the top $N$ feature-target scores, selecting feature-target scores above a given standard deviation $k$, and selection with \gls{mrmr} using these scores.

We restrict our analysis to filter methods because they because they are fit independently from the rest of the training pipeline, and as such do not have a feedback loop from which overfitting could occur \cite{remeseiro2019review}, which are appropriate for small datasets. For all of the aforementioned selection algorithms, as well as the baseline algorithms we compare against, we select $20$ features since it is approximately an eighth of the features. For selection by standard deviations, all features with scores above $\mu + k\sigma$ are selected. We set $k = 0.3$ for all Skipgram and FastText scorers due to low variance, and use $k = 1$ for all others.

\subsection{\Glspl{lm} and Fine-Tuning Datasets} \label{sec:sts_llm}

For generating \gls{sts} scores, we evaluate the performance of several different \gls{llm}, as well as Gensim's \cite{rehurek_lrec} implementation of Skipgram \cite{mikolov2013efficient} and FastText \cite{joulin2016fasttext} for baseline comparisons. We use HuggingFace's \texttt{transformers} \cite{wolf2019huggingface} with the \texttt{sentence\_transformers} library \cite{reimers2019sentence} to fine-tune and produce \gls{sts} scores. We evaluate two models trained on general vocabulary, namely  \texttt{all-MiniLM-L12-v2} \cite{reimers2019sentence}, and \texttt{bert-base-uncased} \cite{devlin2018bert}. We also evaluate the performance of models pre-trained on clinical/scientific text, including Bio\_ClinicalBERT \cite{alsentzer2019publicly}, 
BioMed-RoBERTa-base \cite{biomedroberta}, and PubMedBERT \cite{pubmedbert}.

To fine-tune these models, we train the model to predict cosine similarity on one of two combined \gls{sts} datasets, which we group into a general sentence-level and clinical phrase-level set. We also evaluate the performance of fine-tuning on both sets. Table \ref{tab:ft_vocab} highlights the composition of the selected fine-tuning datsets.

\begin{table*}[h!]
    \centering
    \begin{tabular}{|l|l|l|r|}
        \hline
        \textbf{Dataset} & \textbf{Type} & \textbf{Level} & \textbf{Examples} \\
        \hline
        \texttt{bio\_simlex} \cite{biosimverblex2018} & Clinical & Phrase & 987 \\ 
        \hline
        \texttt{bio\_sim\_verb} \cite{biosimverblex2018} & Clinical & Phrase & 1,000 \\ 
        \hline
        \texttt{mayosrs} \cite{pedersen2007measures} & Clinical & Phrase & 101 \\ 
        \hline
        \textbf{Total Clinical} & & & 2,088 \\
        \hline
        \texttt{sts-companion} \cite{cer-etal-2017-semeval} & General & Sentence & 5,289 \\ 
        \hline
        \texttt{stsb\_multi\_mt;en} \cite{huggingface:dataset:stsb_multi_mt} & General & Sentence & 5,749 \\ 
        \hline
        \textbf{Total General} & & & 11,038 \\
        \hline
        \textbf{Total} & & & 13,126 \\
        \hline
    \end{tabular}
    \caption{Datasets used for training and fine-tuning selected models.}
    \label{tab:ft_vocab}
\end{table*}

ClinicalSTS \cite{xiong2020using} and MedSTS \cite{wang2020medsts} are two clinical sentence-pair datasets, but we were unable to obtain them for fine-tuning. To overcome this limitation, we also evaluate the performance of a Bio\_CinicalBERT model fine-tuned on just the ClinicalSTS dataset \cite{Mulyar_Schumacher_Dredze_2019}.


\section{Experiments}

\subsection{Data}

The data is collected from a partially complete set of participants from the \textit{P5 - Personalized Prediction of Postsurgical Pain} study (IRB \#202101123) conducted at the Washington University/BJC Healthcare System.

A total of 12 surveys were assigned to individual users through the \gls{redcap} system. The principal survey is the one described in \citet{vila2020cognitive}, which contains the four target outcome questions, and are described in more detail in Table \ref{tab:target_questions}. The other surveys include measures of psychological and physical pain, as well as correlated measures.

The dataset was assembled from \gls{redcap} on February 6th, 2023. It contains 1631 responses from a total 617 participants as a part of a final goal of 2,000 participants from the Anonymous Healthcare System. Table \ref{tab:demographics} outlines the key characteristics of the dataset, including number of examples and general demographics.

\begin{table}[h!]
    \centering
    \begin{tabular}{|l|r|}
\hline
 Name                                     &    Value \\
 \hline 
\multicolumn{2}{|c|}{\textbf{\gls{ppsp} Characteristics}}\\
\hline
 Individuals with Complete Mark           & 617      \\
 \hline
 Responses by Individual (mean)   &  2.64        \\
  \hline
 Responses by Individual (std. dev.) & 1.54 \\
 \hline
 Responses by Individual (max)    & 10 \\
 \hline
 \gls{ppsp} (+)                           &  25      \\
 \hline
 \gls{ppsp} (-)                           & 592      \\
 \hline
 \multicolumn{2}{|c|}{\textbf{Race}}\\
 \hline
 Caucasian                                & 497      \\
 \hline
 American Indian / Alaskan Native         &   7      \\
 \hline
 Asian                                    &   4      \\
 \hline
 Black / African Heritage                 &  98      \\
 \hline
 Hawaiian Native / Other Pacific Islander &   1      \\
 \hline
 Other                                    &   9      \\
 \hline
 Prefer not to answer                     &   7      \\
 \hline
 \multicolumn{2}{|c|}{\textbf{Sex Assigned at Birth}}\\
 \hline
 Female                                   & 425      \\
 \hline
 Male                                     & 185      \\
 \hline
 (No Answer)                              & 7       \\
\hline
 \multicolumn{2}{|c|}{\textbf{Age}}\\
 \hline
 Age (min)                                &  19      \\
 \hline
 Age (mean)                               &  52.4686 \\
 \hline
 Age (std. dev.)                          &  13.5219 \\
 \hline
 Age (max)                                &  75      \\
\hline
\end{tabular}
    \caption{Demographics of the partial P5 dataset.}
    \label{tab:demographics}
\end{table}

\subsection{Data \& Model Preparation}

Several steps are taken to prepare the survey data for fitting upon the candidate models.

The first step taken is to prepare the label from this particular survey dataset. The label is derived from the four questions shown in Table \ref{tab:target_questions} is determined using the binary formula $y_1 = (Q_1 \land Q_2) \land (Q_3 \geq 3 \lor Q_4 \geq 3)$. Once the labels are computed, these features are dropped from the dataset. Since we are attempting to predict \gls{ppsp}, we filter out examples that where a column indicating six-month completion has a null value.

\begin{table}[h!]
    \centering
    \begin{tabular}{|l|p{0.7\linewidth}|l|}
        \hline
        \# & Question Text & Type \\
        \hline
        $y_1$ & \texttt{persistent\_pain} & T/F \\
        \hline
        $Q_1$ & In the past week, did you have any pain in your surgical incision or in the area related to your surgery? & Yes/No \\
        \hline
        $Q_2$ & For pain in the area related to your surgery, did the pain start or worsen after the surgery? & Yes/No \\
        \hline
        $Q_3$ & On a scale of zero to ten, with zero being no pain and ten being the worst pain, please fill in your average pain level during the past week, while you were at rest. & 0-10 \\
        \hline
        $Q_4$ & On a scale of zero to ten, with zero being no pain and ten being the worst pain, please fill in your average pain level during the past week, when you were active or moving. & 0-10 \\
        \hline
    \end{tabular}
    \caption{Questions used to determine the label of each survey data example.}
    \label{tab:target_questions}
\end{table}

Participants may fill out the surveys across several responses, and we combine the responses to contain the last reported non-null value for each question. We then select a subset of questions pre-determined to be clinically relevant at any level to \gls{ppsp}, which leaves 131 usable features. Features containing references to image data are then filtered out, and columns containing string-type data with more than $5$ unique values are filtered out. 

To deal with missing entries in survey data, several imputation strategies are applied. For columns with numerical types of data, null entries are replaced with the mean value, and then have the $L_2$ norm applied to that column. Date/time types have the median time imputed, and are then be scaled so the minimum and maximum are 0 and 1 respectively. String types---which we are treating as categorical types given the previous filtering of unique values---will be imputed with the most common value, and then split up into one-hot columns. With these steps, this gives 162 features upon which to train a model. 

Various feature selection and classifier models were tested using the scikit-learn toolkit \cite{pedregosa2011scikit}. Classsifier models tested include XGBoost \cite{chen2016xgboost}, linear \gls{svm}, multilayer perceptron, \gls{gnb}, and \gls{knn}. In addition to testing the proposed variations of \gls{mrmr}, \texttt{SelectFromModel} (which selects based on the weights of a trained model) with linear \gls{svm} and XGBoost are tested. The linear \gls{svm} model is tested with $C$ over 10 logarithmically spaced values from $[10^{-2}, 1]$, while the XGBoost \texttt{SelectFromModel} has the default settings.

An 80\%/20\% train/test split is used for evaluating overall performance, and 5-fold flat \gls{cv} is employed to both select hyperparameters and evaluate the overall performance of the dataset. Nested \gls{cv} is typically employed for evaluating model selection with small datasets, but experimentally may not be necessary with low numbers of hyperparameters while using specific model types, such as gradient boosted trees \cite{wainer2021nested}. For this reason, and due to the fact that nested \gls{cv} with $K$ outer-folds would incur a $K$-fold increase in run-time, flat 5-fold \gls{cv} is used. 
For randomization in NumPy and PyTorch and any of their dependencies, we use the seeds 278797835 and 424989.

\subsection{Feature Selection Implementation}
For baseline selection methods, we employ several filter methods, including selecting the best weights from a linear \gls{svm} model and a XGBoost model, recursive feature elimination, and an implementation of ReliefF by \citet{urbanowicz2018benchmarking}. For ReliefF, we were unable to use 5-fold \gls{cv} due to a bug and only test 10 neighbors for hyperaprameters.

\gls{mi} between features is calculated using the scikit-learn \texttt{mutual\_info\_regression} function, while \gls{mi} between feature and label is calculated using \texttt{mutual\_info\_classif} \cite{pedregosa2011scikit}, which are calculated using the aforementioned \gls{knn} appproach.

Since Skipgram and FastText are initialized from a blank state and require sentence-level examples to evaluate word similarity, we are only able to evaluate the performance of training upon general sentence-pairs for \gls{sts} scoring.

Other than formatting to support the one-hot vectorization of categorical features, the feature names are unmodified. The \texttt{persistent\_pain} target name is a post-hoc addition, but the other target names remain unmodified, and we average their \gls{sts} scores together when evaluating a target. 

\section{Results} \label{chap:results}

\subsection{Overall Performance}

We evaluated 580 different feature selection and classifier pairings. The best performing pairing overall by test \gls{auroc} scores is \gls{mlp} selecting the top $N$ scores, using a linear combination of \gls{mi} scores and \gls{sts} scores from PubMedBERT fine-tuned on the combined vocabulary. We highlight several \gls{mlp} baseline selectors as well as those using PubMedBERT in Table \ref{tab:MLP}.

\begin{table*}
\centering

\begin{tabular}{|l|l|l|l|l|l|l|l|l|}
\hline
\multicolumn{3}{|l|}{Selection Hyperparameters} & \multicolumn{3}{|l|}{AUROC} & \multicolumn{3}{|l|}{AUPRC} \\
\hline
Feature Selector & Scorer & FT Dataset & Test & Train & $\Delta$ & Test & Train & $\Delta$ \\
\hline
Top N$^*$ & MI \& STS & combined & \textbf{0.821} & 0.734 & \textbf{0.087} & \textbf{0.100} & 0.138 & -0.039 \\
Top N$^\dagger$ & STS & combined & 0.736 & 0.737 & \textbf{-0.001} & 0.056 & 0.160 & -0.104 \\
Std. Dev.$^*$$^\dagger$ & MI \& STS & combined & 0.689 & 0.774 & -0.085 & 0.049 & 0.316 & -0.267 \\
mRMR$^*$$^\dagger$ & STS & combined & 0.661 & 0.741 & -0.080 & 0.048 & 0.206 & -0.158 \\
\hline
SFM-LinearSVM & - & - & 0.658 & 0.791 & -0.132 & 0.045 & 0.286 & -0.242 \\
Identity & - & - & 0.576 & \textbf{0.881} & -0.305 & 0.036 & \textbf{0.541} & -0.505 \\
ReliefF & - & - & 0.545 & 0.861 & -0.315 & 0.034 & 0.348 & -0.314 \\
RFE & - & - & 0.534 & 0.834 & -0.299 & 0.034 & 0.452 & -0.418 \\
SFM-XGBoost & - & - & 0.430 & 0.679 & -0.249 & 0.028 & 0.151 & -0.123 \\
mRMR & MI & - & 0.430 & 0.591 & -0.161 & 0.031 & 0.075 & -0.044 \\
Top N & MI & - & 0.331 & 0.521 & -0.190 & 0.023 & 0.048 & -0.025 \\
Std. Dev. & MI & - & 0.259 & 0.494 & -0.235 & 0.021 & 0.044 & \textbf{-0.022} \\
\hline
\end{tabular}
\caption{Selected results for \gls{mlp} with baseline feature selection methods and using PubMedBERT to score features. $*$ is best test \gls{auroc} for given feature selector among ours. $\dagger$ is smallest test-train \gls{auroc} difference for given feature selector among ours.}\label{tab:MLP}
\end{table*}

\subsection{Feature Selector Performance}

\subsubsection{Algorithm Performance} 
\begin{figure*}[h!]
    \centering
    \includegraphics[width=0.9\textwidth]{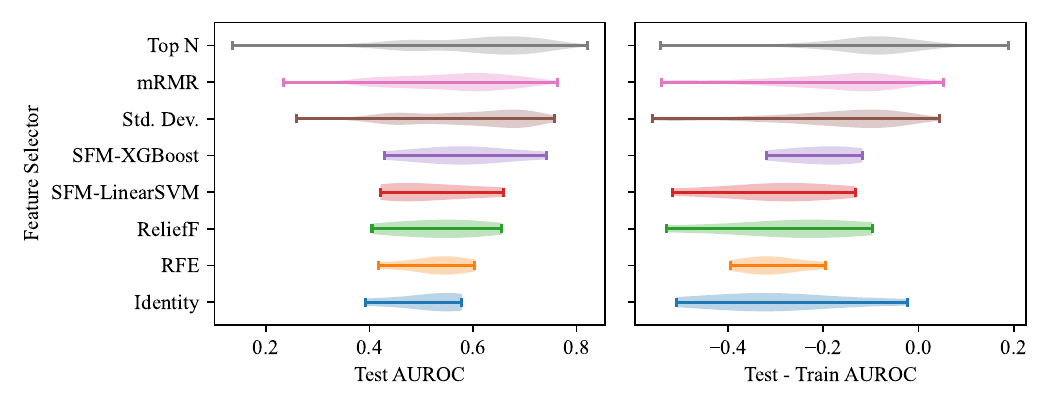}
    \caption{Distribution of feature selector performance by \gls{auroc} scores. Top N, Std. Dev. and \gls{mrmr} include \gls{mi} and \gls{sts} scoring.}
    \label{fig:feature_selector_roc_auc_score_test}
\end{figure*}

As seen in Figure \ref{fig:feature_selector_roc_auc_score_test}, the three selection algorithms that we evaluate with \gls{mi} and \gls{sts} tend to perform better in terms of \gls{auroc} than our baselines. When evaluating the test-train difference, we find that these algorithms tend to have lower levels of overfitting than the baselines.

\subsubsection{Scoring Performance}

\begin{figure*}[h!]
    \centering
    \includegraphics[width=0.9\textwidth]{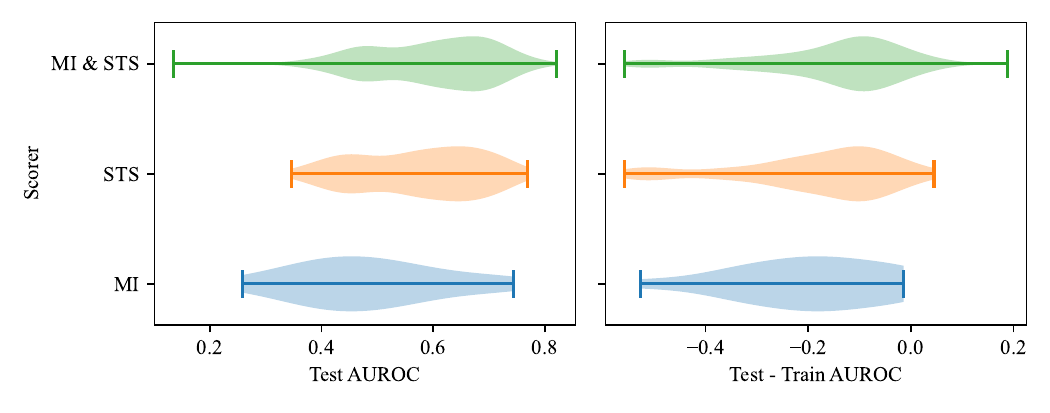}
    \caption{Performance of diferent feature scoring models by \gls{auroc} performance.}
    \label{fig:scorer_roc_auc_score_test}
\end{figure*}

When evaluating the performance between the \gls{sts} and \gls{mi} scorers, as shown in Figure \ref{fig:scorer_roc_auc_score_test}, we find that \gls{sts} tends to perform better at selection than \gls{mi} alone. When using a linear combination of both, we find the distribution of performance appears to be roughly the same as \gls{sts}, but can achieve better test scores. For overfitting performance, we find a similar distribution of behavior. Scoring with \gls{sts} tends to overfit much less than \gls{mi} alone, and adding \gls{mi} does not appear to change the distribution of performance as much. 

For scoring model, shown in Figure \ref{fig:model_roc_auc_score}, we find that the differences between scoring model tend to be smaller. We do note that using a \gls{llm} appears to have a much narrower distribution of performance than the \glspl{lm} tested, and the differences between choice of \gls{llm} appears to be smaller.

\begin{figure}[h!]
    \centering
    \includegraphics[width=0.45\textwidth]{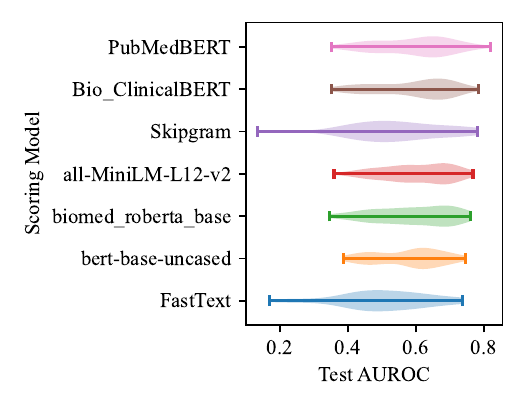}
    \caption{Distribution of \gls{sts} scoring model performance by test \gls{auroc}}
    \label{fig:model_roc_auc_score}
\end{figure}

\begin{figure*}[h!]
    \centering
    \begin{subfigure}[b]{0.49\textwidth}
         \centering
        \includegraphics[width=\textwidth]{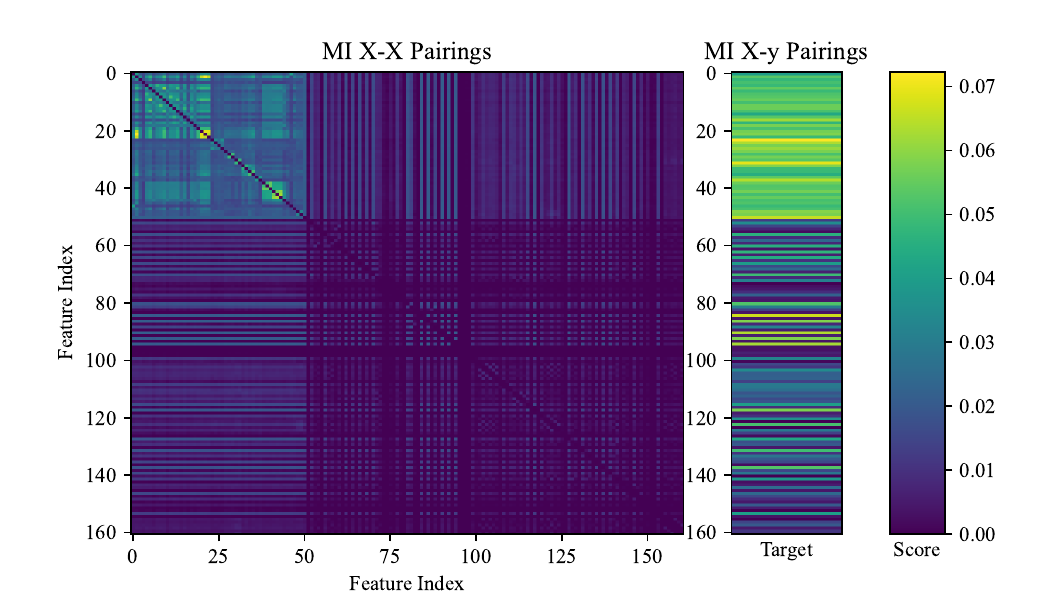}
        \caption{\gls{mi} scoring}
        \label{fig:mi_heatmap}
    \end{subfigure}
    \\
    \begin{subfigure}[b]{0.49\textwidth}
        \centering
        \includegraphics[width=\textwidth]{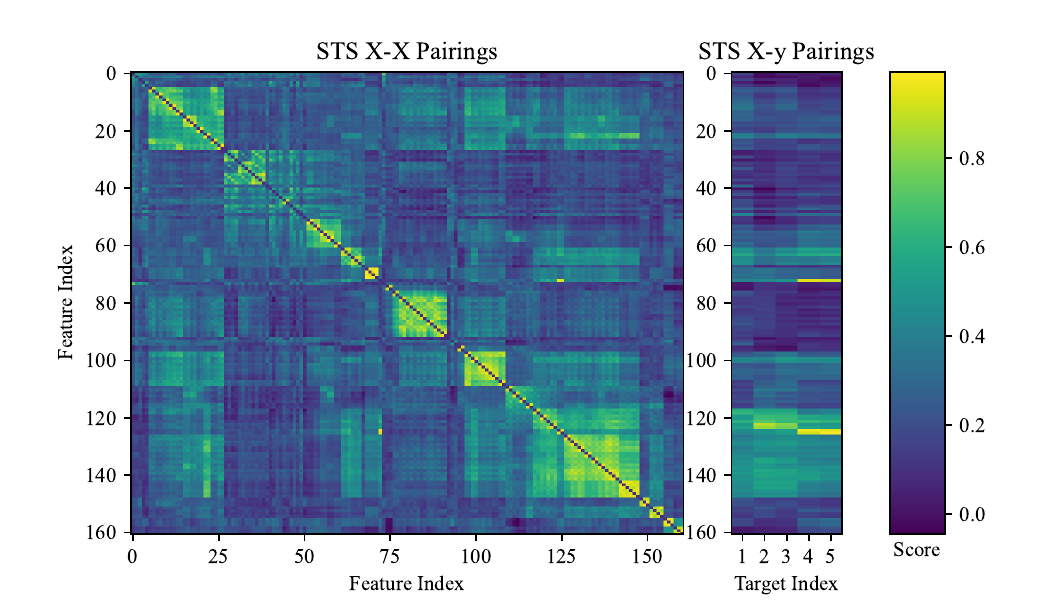}
        \caption{\gls{sts} scoring with PubMedBERT fine-tuned on the general vocab pairings.}
        \label{fig:sts_heatmap_gen}
    \end{subfigure}
    \hfill
    \begin{subfigure}[b]{0.49\textwidth}
        \centering
        \includegraphics[width=\textwidth]{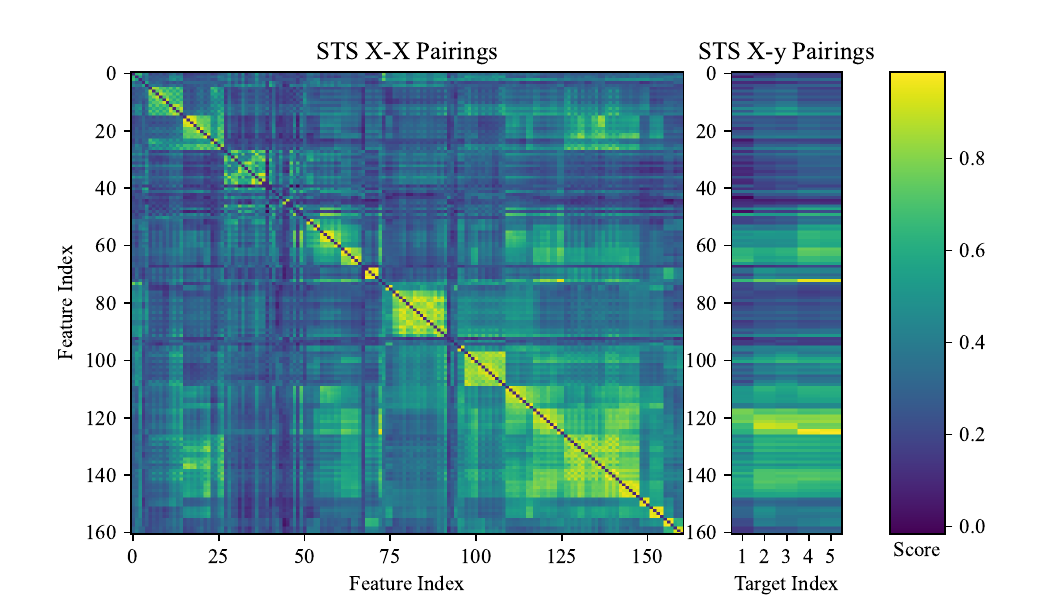}
        \caption{\gls{sts} scoring with PubMedBERT fine-tuned on the clinical vocab pairings.}
        \label{fig:sts_heatmap_clin}
    \end{subfigure}

    \caption{Similarity pairings for different scoring models. Feature pairings are mirrored across the diagonal for viewability.}
    \label{fig:mi_vs_sts_pairings}
\end{figure*}

Finally, we find that the differences between a clinical and general fine-tuning dataset, shown in Figure \ref{fig:model_roc_auc_score}, appear to negligible overall. This does not mean, however, that fine-tuning dataset has no overall effect. The best clinical/scientific pre-trained model, PubMedBERT, performed best with the combined fine-tuning dataset on  \gls{mlp}, which we believe is partially attributable to the difference between those scores and scores produced from the general dataset, as seen in Figures \ref{fig:sts_heatmap_clin} and \ref{fig:sts_heatmap_gen}, respectively.

\begin{figure}[h!]
    \centering
    \includegraphics[width=0.45\textwidth]{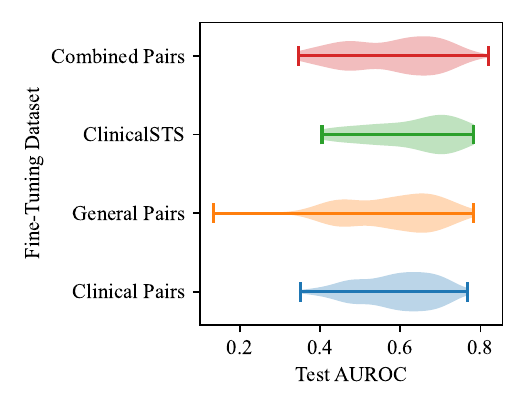}
    \caption{Distribution of \gls{sts} fine-tuning dataset performance by test \gls{auroc}}
    \label{fig:model_roc_auc_score}
\end{figure}

\subsubsection{Selected Features}

When evaluating the selected features, we see a noticeable difference between those selected using \gls{mi} and \gls{sts} scores. Figure \ref{fig:mi_vs_sts_pairings}, which shows \gls{mi} and selected \gls{sts} scores for the features and targets for the dataset, highlights why this difference exists, as the \gls{sts} scores are capable of highlighting relationships between features and targets. The differences in the features selected, are shown in the supplemental material, as well as the performance of the model using the \gls{shap} \cite{lundberg2017unified} model when used with \acrlong{gnb}.

\section{Discussion}
\subsection{Feature Selection Performance}

As shown in the previous section, using \gls{sts} to select features appears to offer significant performance benefits over traditional feature selection methods.

One principal reason that the usage of \gls{sts} appears to work better than statistical measures like \gls{mi} is that \gls{sts} scores are not affected by the curse of dimensionality. As can be seen with \gls{mi}, many overlapping relationships can cause individual features to share little information, and features that we would expect to be more relevant than others will only have slightly more \gls{mi} than those that would not be relevant. This is particularly evident when looking at \gls{mrmr} scores calculated with \gls{mi} scores, seen in the supplementary material, that dip into the negative: the subsequent feature learned is more redundant than it is relevant, yet there are many unselected features that we would consider to be truly relevant.

\Gls{sts} also provides a contrasting, non-correlated, source of information compared to statistical measures like \gls{mi}. This fact can be seen from the strong contrast in highlighted regions between Figures \ref{fig:mi_heatmap} and \ref{fig:sts_heatmap_clin}/\ref{fig:sts_heatmap_gen}. When used in \gls{mrmr}, \gls{sts} is clearly more able to highlight redundant regions, especially along the diagonal, and is more able to distinguish between relevant and irrelevant features than \gls{mi}. This is particularly useful for when \gls{mi} that results from the training dataset fails to match what we might expect from the population sampled, as \gls{sts} is not vulnerable to differences in the sample and population distributions.

\subsection{Survey Writing \& Feature Naming}

The surveys used here were not designed in anticipation of this type of feature selection algorithm, but we anticipate survey designers may want to write questions with consideration to \gls{sts}-based feature selection. We offer several guidelines for writing surveys or creating other forms of data with descriptive feature names.

Considerations should be made with respect to vocabulary. Words should primarily be derived from the vocabulary used to create tokens, and secondarily from the vocabulary used to fine-tune the pre-trained model to produce \gls{sts} scores. Even if a ground-truth label is not given for a possible word-level \gls{sts} pairing, the \gls{lm} should be able to produce a meaningful \gls{sts} score, as the \gls{lm} will have learned word relationships in its vocabulary during pre-training. Acronyms should be expanded where possible, as an acronym may not be in either vocabulary, and thus be semantically meaningless.

Although the evidence presented here suggests that there may be a negligible difference between the performance of tokenizers---as the models tested use different tokenizers---subtle differences in tokenization may change the way that semantic relationships are captured. For example, \citet{bostrom2020byte} note that byte-pair encoding \cite{gage1994bpe, sennrich-etal-2016-neural} has weaker performance than unigram language modeling \cite{kudo-2018-subword} with respect to morphological segmentations, implying that features tokenized with the latter algorithm may produce more meaningful \gls{sts} scores.

\subsection{Future Work}

One area of future work is evaluating the performance of other measures, as the scorers evaluated here both have limitations. \gls{mi} is incapable of measuring the true amount of information a feature contains in context with other features, and \gls{sts} only represent semantic relationships without any regard to their underlying statistical relationship.

Future work could also consider the choice of other models for computing \gls{sts}. We were not able to perform a complete evaluation due to resource limitations, and further analyses could evaluate larger models such as BioGPT \cite{luo2022biogpt} and GatorTron \cite{yang2022large}. The evidence presented here suggests that there may be marginal benefits to different scoring models.

Future work regarding \gls{sts}-based feature selection could also consider the use of semantic pairs that specifically rate \textit{relevancy} and \textit{redundancy} between pairs of question embeddings, rather than similarity. Relevant and redundant questions may not always be semantically similar, and a dataset for fine-tuning upon this type of task may improve the performance of feature selection techniques that use \gls{sts}.

Another potential area of future work would be to serialize the the feature selection task into a text prompt. Serialization of tablular data into a question prompt for a large language model can achieve high performance in a few-shot learning context \cite{hegselmann2022tabllm}, and serialization of a feature selection objective may also be able to capture further semantic relationships between features.

\subsection{Acknowledgements}
This work was made possible through the support of the Department of Defense Congressionally Directed Medical Research Programs award W81XWH-21-1-0736. The authors would like to also thank Hanyang Liu for his valuable guidance on the final writing and publishing of this project.
\bigskip

\bibliography{cite}

\end{document}


\maketitle

\tableofcontents

\pagebreak

\section{Feature Importance By Selection Algorithm}

This section contains the results for each of our selection algorithms that use \gls{mi}, \gls{sts}, and a linear combination of both. In all cases, we generate \gls{sts} with PubMedBERT fine-tuned on the combined vocabulary. In addition, we provide \gls{shap} values on a \gls{mlp} model trained with the filtered features.

\subsection{Top 20 Scores, \acrshort{mi}}
\begin{longtable}{|p{0.3\textwidth}|l|p{0.135\textwidth}|l|l|p{0.125\textwidth}|}
\hline

 Feature Name & $n$ & Relevancy & Mean Absolute \acrshort{shap} \\ \hline

 Current Medications:  Confirmed with participant (choice=Metformin)\_Checked                                                                                          &       1 & 0.0722903 & 1.37293772e-09 \\
 \hline
 Color Score (C), Raw Score                                                                                                                                           &       2 & 0.0696211 & 1.14889901e-07 \\
 \hline
 Please specify your race: (choice=Black / African Heritage)\_Checked                                                                                                  &       3 & 0.0668161 & 0                         \\
 \hline
 BRS Score                                                                                                                                                            &       4 & 0.0660192 & 0                         \\
 \hline
 Color-Word Score (CW), Residual                                                                                                                                      &       5 & 0.0639546 & 0                         \\
 \hline
 Please specify your race: (choice=Prefer not to answer)\_Checked                                                                                                      &       6 & 0.0638041 & 0                         \\
 \hline
 Current Medications:  Confirmed with participant (choice=Metformin)\_Unchecked                                                                                        &       7 & 0.0634184 & 0                         \\
 \hline
 Other:.1\_married; polyam                                                                                                                                             &       8 & 0.062555  & 0                         \\
 \hline
 Current Medications:  Confirmed with participant (choice=SNRI antidepressants (venlafaxine, duloxetine, desvenlafaxine, milnacipran))\_Unchecked                      &       9 & 0.0619615 & 0                         \\
 \hline
 Current Medications:  Confirmed with participant (choice=None of the above)\_Unchecked                                                                                &      10 & 0.0611848 & 0                         \\
 \hline
 Magnification Score                                                                                                                                                  &      11 & 0.0591411 & 0                         \\
 \hline
 Color Score (C), Age/Ed. Predicted                                                                                                                                   &      12 & 0.0588759 & 7.82877036e-09 \\
 \hline
 Please specify your race: (choice=Hawaiian Native / Other Pacific Islander)\_Checked                                                                                  &      13 & 0.0584943 & 0                         \\
 \hline
 Current Medications:  Confirmed with participant (choice=Opioids (morphine, oxycodone, fentanyl, tramadol, methadone, buprenorphine))\_Checked                        &      14 & 0.0583059 & 0                         \\
 \hline
 Total Years of Education                                                                                                                                             &      15 & 0.0581167 & 3.84597352e-09    \\
 \hline
 Have you ever been diagnosed with chronic pain?\_No                                                                                                                   &      16 & 0.0579427 & 0                         \\
 \hline
 Helplessness Score                                                                                                                                                   &      17 & 0.0578844 & 2.42765435e-06 \\
 \hline
 Current Medications:  Confirmed with participant (choice=None of the above)\_Checked                                                                                  &      18 & 0.0577642 & 0                         \\
 \hline
 Word Score (W), Raw Score                                                                                                                                            &      19 & 0.0576109 & 2.98490162e-07 \\
 \hline
 Current Medications:  Confirmed with participant (choice=SSRI antidepressants (citalopram, escitalopram, fluoxetine, fluvoxamine, paroxetine, sertraline))\_Unchecked &      20 & 0.0574527 & 0                         \\
\hline

\caption{Feature importance using \acrshort{mlp} using \gls{mi} selecting 20 features. }
\label{tab:mrmr_mlp_linear}

\end{longtable}

\subsection{Top 20 Scores, \acrshort{sts}}
\begin{longtable}{|p{0.3\textwidth}|l|p{0.135\textwidth}|l|l|p{0.125\textwidth}|}
\hline
 Feature Name & $n$ & Relevancy & Mean Absolute \acrshort{shap} \\ 
\hline
 On a scale of zero to ten, with zero being no pain and ten being the worst pain, please mark your average pain level during the past week, while at rest.                      &       1 & 0.645585 &  1.46336854e-06 \\
 \hline
 On a scale of zero to ten, with zero being no pain and ten being the  worst pain , please mark your  average pain level during the past week, while you were active or moving. &       2 & 0.645342 &  4.46294155e-07 \\
 \hline
 Have you been experiencing pain in the past week?\_No                                                                                                                           &       3 & 0.639949 &  0.00279322         \\
 \hline
 Have you been experiencing pain in the past week?\_Yes, and RELATED to my need for surgery                                                                                      &       4 & 0.638486 &  0.00297259         \\
 \hline
 Have you been experiencing pain in the past week?\_Yes, but NOT RELATED to my need for surgery                                                                                  &       5 & 0.634165 &  0.00894911         \\
 \hline
 On a scale of zero to ten, with zero being no pain and ten being the worst pain, what is your average pain level at rest TODAY?                                                &       6 & 0.619892 & 4.25537234e-07 \\
 \hline
 In the past three months, did you experience daily or near daily pain?\_No                                                                                                      &       7 & 0.544496 &  0.01075854         \\
 \hline
 Have you ever been diagnosed with chronic pain?\_No                                                                                                                             &       8 & 0.534078 &  0.01045607         \\
 \hline
 Have you ever been diagnosed with chronic pain?\_Yes                                                                                                                            &       9 & 0.531654 &  0.00238557         \\
 \hline
 In the past three months, did you experience daily or near daily pain?\_Yes                                                                                                     &      10 & 0.526172 &  0.00067493         \\
 \hline
 Pain intensity - Baseline Average:                                                                                                                                             &      11 & 0.525021 &  1.61967588e-06 \\
 \hline
 Pain intensity - Baseline 1:                                                                                                                                                   &      12 & 0.487709 &  2.1284362e-06   \\
 \hline
 Do you take any of the following medications for pain treatment at least once per week? (choice=None)\_Unchecked                                                                &      13 & 0.478695 &  0.00369703         \\
 \hline
 Pain intensity - Baseline 2:                                                                                                                                                   &      14 & 0.475303 &  1.34237724e-06 \\
 \hline
 Do you take any of the following medications for pain treatment at least once per week? (choice=Other)\_Unchecked                                                               &      15 & 0.475231 &  0.00833284         \\
 \hline
 Do you take any of the following medications for pain treatment at least once per week? (choice=None)\_Checked                                                                  &      16 & 0.461651 &  0.00497855         \\
 \hline
 Do you take any of the following medications for pain treatment at least once per week? (choice=Duloxetine (Cymbalta) or amitriptyline (Elavil))\_Unchecked                     &      17 & 0.449812 &  0.0008251           \\
 \hline
 Have you used opioid medications for pain management in the past?\_I don't know                                                                                                 &      18 & 0.44269  &  0.00038904         \\
 \hline
 Have you used opioid medications for pain management in the past?\_yes                                                                                                          &      19 & 0.43422  &  0.00495188         \\
 \hline
 Do you take any of the following medications for pain treatment at least once per week? (choice=Other)\_Checked                                                                 &      20 & 0.430055 &  0.00134032         \\
 \hline
\caption{Feature importance using \acrshort{mlp} using \gls{sts} scoring with PubMedBERT fine-tuned on the combined vocab selecting 20 features. }
\label{tab:mrmr_mlp_linear}

\end{longtable}

\subsection{Top 20 Scores, \acrshort{mi} \& \acrshort{sts}}
\begin{longtable}{|p{0.3\textwidth}|l|p{0.135\textwidth}|l|l|p{0.125\textwidth}|}
\hline
 Feature Name & $n$ & Relevancy & Mean Absolute \acrshort{shap} \\ \hline

 Have you ever been diagnosed with chronic pain?\_No                                                                                                         &       1 & 0.734467 & 0.01120194         \\
 \hline
 Have you ever been diagnosed with chronic pain?\_Yes                                                                                                        &       2 & 0.712339 & 0.00546406         \\
 \hline
 Have you been experiencing pain in the past week?\_No                                                                                                       &       3 & 0.599768 & 0.00223598         \\
 \hline
 Have you been experiencing pain in the past week?\_Yes, and RELATED to my need for surgery                                                                  &       4 & 0.598852 & 5.00820274e-07 \\
 \hline
 Have you been experiencing pain in the past week?\_Yes, but NOT RELATED to my need for surgery                                                              &       5 & 0.529267 & 1.79035372e-06 \\
 \hline
 In the past three months, did you experience daily or near daily pain?\_Yes                                                                                 &       6 & 0.525755 & 0.00294529         \\
 \hline
 Do you take any of the following medications for pain treatment at least once per week? (choice=Duloxetine (Cymbalta) or amitriptyline (Elavil))\_Unchecked &       7 & 0.512279 & 0.00100577         \\
 \hline
 In the past three months, did you experience daily or near daily pain?\_No                                                                                  &       8 & 0.504445 & 0.00930347         \\
 \hline
 Helplessness Score                                                                                                                                         &       9 & 0.493851 & 5.01834836e-07 \\
 \hline
 Pain intensity - Baseline 2:                                                                                                                               &      10 & 0.479731 & 9.24779374e-07 \\
 \hline
 Pain intensity - Baseline Average:                                                                                                                         &      11 & 0.478815 & 5.96666072e-07 \\
 \hline
 Do you take any of the following medications for pain treatment at least once per week? (choice=Other)\_Unchecked                                           &      12 & 0.472314 &  0.00909019    \\
 Have you used opioid medications for pain management in the past?\_yes                                                                                      &      13 & 0.466008 & 0.00462665         \\
 \hline
 Pain intensity - CPM 1:                                                                                                                                    &      14 & 0.457419 & 4.55048919e-07 \\
 \hline
 Do you take any of the following medications for pain treatment at least once per week? (choice=None)\_Unchecked                                            &      15 & 0.455803 & 0.00129691         \\
 \hline
 Pain intensity - CPM 2:                                                                                                                                    &      16 & 0.444577 & 1.87985214e-06 \\
 \hline
 Pain intensity - CPM Average:                                                                                                                              &      17 & 0.439207 & 4.87195489e-07 \\
 \hline
 Do you take any of the following medications for pain treatment at least once per week? (choice=Tramadol (Ultram))\_Unchecked                               &      18 & 0.438722 & 0.00499209         \\
 \hline
 Have you used opioid medications for pain management in the past?\_I don't know                                                                             &      19 & 0.43665  & 0.0006287           \\
 \hline
 Have you used opioid medications for pain management in the past?\_no                                                                                       &      20 & 0.433569 & 3.60313277e-05 \\
\hline

\caption{Feature importance using \acrshort{mlp} using linear scoring with PubMedBERT fine-tuned on the combined vocab selecting 20 features. }
\label{tab:mrmr_mlp_linear}

\end{longtable}

\subsection{\acrshort{mrmr}, \acrshort{mi}}
\begin{longtable}{|p{0.3\textwidth}|l|p{0.135\textwidth}|l|l|p{0.125\textwidth}|}
\hline

 Feature Name & $n$ & mRMR Score & Relevancy & Redundancy & Mean Absolute \acrshort{shap} \\ \hline

Current Medications:  Confirmed with participant (choice=Metformin)\_Checked                                                                                  &       1 &       0.0723 &      0.0723 &       0      & 1.80786083e-07 \\
\hline
 Please specify your race: (choice=Black / African Heritage)\_Checked                                                                                          &       2 &       0.0668 &      0.0668 &       0      & 1.31254826e-07 \\
 \hline
 Please specify your race: (choice=Hawaiian Native / Other Pacific Islander)\_Checked                                                                          &       3 &       0.0585 &      0.0585 &       0      & 2.02387005e-07 \\
 \hline
 Please specify your race: (choice=American Indian / Alaskan Native)\_Checked                                                                                  &       4 &       0.0523 &      0.0523 &       0      & 1.2439402e-07   \\
 \hline
 Baseline Emotional Distress-Anxiety - Short Form 4a T Score                                                                                                  &       5 &       0.0461 &      0.0461 &       0      & 2.17081369e-07 \\
 \hline
 Baseline Cognitive Function - Abilities - Short Form 4a T Score                                                                                              &       6 &       0.0119 &      0.0352 &       0.0232 & 1.46263805e-07 \\
 \hline
 Please specify your ethnicity:\_Non-Hispanic                                                                                                                  &       7 &      -0.0011 &      0.0227 &       0.0238 & 0.00047083         \\
 \hline
 Did you find submerging your hand in the cold water bath to be:\_Not Painful                                                                                  &       8 &      -0.0049 &      0.0497 &       0.0545 & 3.61855009e-07 \\
 \hline
 Pain intensity - Baseline 2:                                                                                                                                 &       9 &      -0.0145 &      0.0474 &       0.0619 & 3.04670314e-07 \\
 \hline
 Pain intensity - CPM Average:                                                                                                                                &      10 &      -0.012  &      0.04   &       0.052  & 3.40698218e-07 \\
 \hline
 Baseline Emotional Distress-Depression - Short Form 4a T Score                                                                                               &      11 &      -0.016  &      0.024  &       0.04   & 4.35499288e-07 \\
 \hline
 How often have you been using cannabis (marijuana, pot, weed, grass) in the past year?\_Never                                                                 &      12 &      -0.0286 &      0.0408 &       0.0694 & 0.00359785         \\
 \hline
 Pain intensity - CPM 1:                                                                                                                                      &      13 &      -0.0479 &      0.0446 &       0.0924 & 2.01787643e-07 \\
 \hline
 Moderate activities such as moving a table, pushing a vacuum cleaner, bowling, or playing golf?\_Yes limited a lot                                            &      14 &      -0.0585 &      0.0203 &       0.0789 & 0.00129293         \\
 \hline
 Climbing several flights of stairs?\_Yes limited a lot                                                                                                        &      15 &      -0.0591 &      0.011  &       0.0701 & 0.00302154         \\
 \hline
 Why are you having this upcoming procedure? Please mark all that apply. Please mark all that apply. (choice=My doctor said I needed the procedure)\_Unchecked &      16 &      -0.0536 &      0.0293 &       0.0829 & 0.00573359         \\
 \hline
 Do you take any of the following medications for pain treatment at least once per week? (choice=Tramadol (Ultram))\_Unchecked                                 &      17 &      -0.0511 &      0.0502 &       0.1013 & 0.00229605         \\
 \hline
 Did you find submerging your hand in the cold water bath to be:\_Mildly Painful                                                                               &      18 &      -0.0524 &      0.025  &       0.0773 & 6.09845743e-07 \\
 \hline
 Have you ever been diagnosed with anxiety?\_No                                                                                                                &      19 &      -0.0599 &      0.0289 &       0.0888 & 1.61796364e-06 \\
 \hline
 Have you ever been diagnosed with anxiety?\_Yes                                                                                                               &      20 &      -0.0611 &      0.0263 &       0.0874 & 3.27162909e-07 \\
 \hline

    \caption{Feature importance using \acrshort{mlp} using \acrshort{mrmr} with \acrshort{mi} scores with 20 features}
    \label{tab:mb_feature_importance}

\end{longtable}

\subsection{\acrshort{mrmr}, \acrshort{sts}}
\begin{longtable}{|p{0.3\textwidth}|l|p{0.135\textwidth}|l|l|p{0.125\textwidth}|}
\hline

 Feature Name & $n$ & \acrshort{mrmr} & Relevancy & Redundancy & Mean Absolute \acrshort{shap} \\ \hline
On a scale of zero to ten, with zero being no pain and ten being the worst pain, please mark your average pain level during the past week, while at rest.                      &       1 &       0.6456 &      0.6456 &       0      & 1.43756747e-06 \\
\hline
 On a scale of zero to ten, with zero being no pain and ten being the  worst pain , please mark your  average pain level during the past week, while you were active or moving. &       2 &       0.6453 &      0.6453 &       0      & 1.76442483e-06 \\
 \hline
 Do you take any of the following medications for pain treatment at least once per week? (choice=Aspirin)\_Unchecked                                                             &       3 &       0.2493 &      0.397  &       0.1477 & 0.00140821         \\
 \hline
 Do you take any of the following medications for pain treatment at least once per week? (choice=Tramadol (Ultram))\_Checked                                                     &       4 &       0.1669 &      0.3962 &       0.2294 & 0.00218598         \\
 \hline
 Have you ever been diagnosed with chronic pain?\_No                                                                                                                             &       5 &       0.1206 &      0.5341 &       0.4135 & 0.01126774         \\
 \hline
 Have you ever been diagnosed with chronic pain?\_Yes                                                                                                                            &       6 &       0.2025 &      0.5317 &       0.3291 & 0.00169776         \\
 \hline
 In the past three months, did you experience daily or near daily pain?\_Yes                                                                                                     &       7 &       0.1023 &      0.5262 &       0.4239 & 0.00016405         \\
 \hline
 Have you been experiencing pain in the past week?\_Yes, but NOT RELATED to my need for surgery                                                                                  &       8 &       0.1518 &      0.6342 &       0.4823 & 0.00909981         \\
 \hline
 On a scale of zero to ten, with zero being no pain and ten being the worst pain, what is your average pain level at rest TODAY?                                                &       9 &       0.114  &      0.6199 &       0.5059 & 2.0828322e-07   \\
 \hline
 Have you been experiencing pain in the past week?\_No                                                                                                                           &      10 &       0.0777 &      0.6399 &       0.5622 & 0.00350571         \\
 \hline
 Rumination Score                                                                                                                                                               &      11 &       0.0543 &      0.2666 &       0.2123 & 2.4462886e-06   \\
 \hline
 Have you been experiencing pain in the past week?\_Yes, and RELATED to my need for surgery                                                                                      &      12 &       0.0627 &      0.6385 &       0.5758 & 0.0041694           \\
 \hline
 Helplessness Score                                                                                                                                                             &      13 &       0.0643 &      0.3075 &       0.2433 & 1.75271276e-06 \\
 \hline
 Pain intensity - Baseline Average:                                                                                                                                             &      14 &       0.0722 &      0.525  &       0.4528 & 2.71861889e-06 \\
 In the past three months, did you experience daily or near daily pain?\_No                                                                                                      &      15 &       0.0952 &      0.5445 &       0.4493 & 0.00994588         \\
 \hline
 Pain intensity - Baseline 1:                                                                                                                                                   &      16 &       0.0679 &      0.4877 &       0.4199 & 1.89465065e-07 \\
 \hline
 Pain intensity - Baseline 2:                                                                                                                                                   &      17 &       0.0749 &      0.4753 &       0.4004 & 1.72906497e-06 \\
 \hline
 Why are you having this upcoming procedure? Please mark all that apply. Please mark all that apply. (choice=Decrease pain)\_Unchecked                                           &      18 &       0.0809 &      0.3686 &       0.2877 & 2.52886547e-06 \\
 \hline
 Why are you having this upcoming procedure? Please mark all that apply. Please mark all that apply. (choice=Decrease pain)\_Checked                                             &      19 &       0.0727 &      0.3482 &       0.2756 & 0                         \\
 \hline
 Do you take any of the following medications for pain treatment at least once per week? (choice=Tramadol (Ultram))\_Unchecked                                                   &      20 &       0.0795 &      0.4162 &       0.3368 & 0.00644815         \\
 \hline

\caption{Feature importance using \acrshort{mlp} using \acrshort{mrmr} with \acrshort{sts} scores from PubMedBERT fine-tuned on the combined vocab with 20 features.}
\label{tab:mqo_feature_importance}

\end{longtable}

\subsection{\acrshort{mrmr}, \acrshort{mi} \& \acrshort{sts}}

\begin{longtable}{|p{0.3\textwidth}|l|p{0.135\textwidth}|l|l|p{0.125\textwidth}|}
\hline
 Feature Name & $n$ & \acrshort{mrmr} & Relevancy & Redundancy & Mean Absolute \acrshort{shap} \\ \hline
 Have you ever been diagnosed with chronic pain?\_No                                                                                                         &       1 &       0.7345 &      0.7345 &       0      & 0.01284267         \\
 \hline
 Have you ever been diagnosed with chronic pain?\_Yes                                                                                                        &       2 &       0.7123 &      0.7123 &       0      & 0.00255372         \\
 \hline
 Baseline Emotional Distress-Anxiety - Short Form 4a                                                                                                        &       3 &       0.1982 &      0.3735 &       0.1753 & 2.61222337e-05 \\
 \hline
 In the past three months, did you experience daily or near daily pain?\_Yes                                                                                 &       4 &       0.158  &      0.5258 &       0.3677 & 0.00154897         \\
 \hline
 Do you take any of the following medications for pain treatment at least once per week? (choice=Duloxetine (Cymbalta) or amitriptyline (Elavil))\_Unchecked &       5 &       0.1109 &      0.5123 &       0.4013 & 0.00176415         \\
 \hline
 Baseline Emotional Distress-Anxiety - Short Form 4a T Score                                                                                                &       6 &       0.1008 &      0.3454 &       0.2446 & 1.33594353e-05 \\
 \hline
 Have you used opioid medications for pain management in the past?\_I don't know                                                                             &       7 &       0.0502 &      0.4366 &       0.3865 & 0.00153403         \\
 \hline
 Have you been experiencing pain in the past week?\_No                                                                                                       &       8 &       0.0596 &      0.5998 &       0.5402 & 0.00508954         \\
 \hline
 Please specify your ethnicity:\_Hispanic                                                                                                                    &       9 &       0.0652 &      0.2196 &       0.1545 & 1.19111503e-05 \\
 \hline
 On a scale of zero to ten, with zero being no pain and ten being the worst pain, what is your average pain level at rest TODAY?                            &      10 &       0.0667 &      0.3974 &       0.3307 & 1.6230635e-05   \\
 \hline
 Have you been experiencing pain in the past week?\_Yes, and RELATED to my need for surgery                                                                  &      11 &       0.0858 &      0.5989 &       0.513  & 2.27270858e-05 \\
 \hline
 Pain intensity - Baseline 2:                                                                                                                               &      12 &       0.0716 &      0.4797 &       0.4081 & 1.82589818e-05 \\
 \hline
 In the past three months, did you experience daily or near daily pain?\_No                                                                                  &      13 &       0.0678 &      0.5044 &       0.4366 & 0.01732398         \\
 \hline
 Other:.1\_married; polyam                                                                                                                                   &      14 &       0.0524 &      0.3322 &       0.2798 & 1.78507086e-05 \\
 \hline
 Pain intensity - Baseline 1:                                                                                                                               &      15 &       0.0609 &      0.3951 &       0.3342 & 3.09535235e-05 \\
 \hline
 Why are you having this upcoming procedure? Please mark all that apply. Please mark all that apply. (choice=Decrease pain)\_Unchecked                       &      16 &       0.0512 &      0.3042 &       0.253  & 1.89781409e-05 \\
 \hline
 Pain intensity - Baseline Average:                                                                                                                         &      17 &       0.0574 &      0.4788 &       0.4214 & 2.68615232e-05 \\
 \hline
 Have you been experiencing pain in the past week?\_Yes, but NOT RELATED to my need for surgery                                                              &      18 &       0.0602 &      0.5293 &       0.4691 & 2.64203743e-05 \\
 \hline
 Pain intensity - CPM 2:                                                                                                                                    &      19 &       0.0499 &      0.4446 &       0.3947 & 1.49991297e-05 \\
 \hline
 Pain intensity - CPM 1:                                                                                                                                    &      20 &       0.0372 &      0.4574 &       0.4203 & 2.22239174e-05 \\
 \hline

\caption{Feature importance using \acrshort{mlp} using linear scoring with PubMedBERT fine-tuned on the combined vocab selecting 20 features. }
\label{tab:mrmr_mlp_linear}

\end{longtable}

\subsection{$\mu + 1\sigma$, \acrshort{sts}}

\begin{longtable}{|p{0.3\textwidth}|l|p{0.135\textwidth}|l|l|p{0.125\textwidth}|}
\hline
 Feature Name & $n$ & Score & Mean Absolute \acrshort{shap} \\ \hline

  Current Medications:  Confirmed with participant (choice=Metformin)\_Checked                                                                                          &       1 & 0.0722903 &  0   \\
  \hline
 Color Score (C), Raw Score                                                                                                                                           &       2 & 0.0696211 & 9.05460563e-06    \\
 \hline
 Please specify your race: (choice=Black / African Heritage)\_Checked                                                                                                  &       3 & 0.0668161 & 7.03006661e-10    \\
 \hline
 BRS Score                                                                                                                                                            &       4 & 0.0660192 & 9.07897635e-09    \\
 \hline
 Color-Word Score (CW), Residual                                                                                                                                      &       5 & 0.0639546 & 1.20770991e-08    \\
 \hline
 Please specify your race: (choice=Prefer not to answer)\_Checked                                                                                                      &       6 & 0.0638041 & 4.17793688e-09    \\
 \hline
 Current Medications:  Confirmed with participant (choice=Metformin)\_Unchecked                                                                                        &       7 & 0.0634184 & 1.18285792e-09    \\
 \hline
 Other:.1\_married; polyam                                                                                                                                             &       8 & 0.062555  & 0   \\
 \hline
 Current Medications:  Confirmed with participant (choice=SNRI antidepressants (venlafaxine, duloxetine, desvenlafaxine, milnacipran))\_Unchecked                      &       9 & 0.0619615 & 0   \\
 \hline
 Current Medications:  Confirmed with participant (choice=None of the above)\_Unchecked                                                                                &      10 & 0.0611848 & 0   \\
 \hline
 Magnification Score                                                                                                                                                  &      11 & 0.0591411 & 3.05139839e-09    \\
 \hline
 Color Score (C), Age/Ed. Predicted                                                                                                                                   &      12 & 0.0588759 & 2.06625958e-08    \\
 \hline
 Please specify your race: (choice=Hawaiian Native / Other Pacific Islander)\_Checked                                                                                  &      13 & 0.0584943 & 0   \\
 \hline
 Current Medications:  Confirmed with participant (choice=Opioids (morphine, oxycodone, fentanyl, tramadol, methadone, buprenorphine))\_Checked                        &      14 & 0.0583059 & 0   \\
 \hline
 Total Years of Education                                                                                                                                             &      15 & 0.0581167 & 5.38566368e-09    \\
 \hline
 Have you ever been diagnosed with chronic pain?\_No                                                                                                                   &      16 & 0.0579427 & 3.60115734e-10    \\
 \hline
 Helplessness Score                                                                                                                                                   &      17 & 0.0578844 & 8.02095497e-06    \\
 \hline
 Current Medications:  Confirmed with participant (choice=None of the above)\_Checked                                                                                  &      18 & 0.0577642 & 0   \\
 \hline
 Word Score (W), Raw Score                                                                                                                                            &      19 & 0.0576109 & 1.09118178e-05    \\
 \hline
 Current Medications:  Confirmed with participant (choice=SSRI antidepressants (citalopram, escitalopram, fluoxetine, fluvoxamine, paroxetine, sertraline))\_Unchecked &      20 & 0.0574527 & 4.59364778e-10    \\
 \hline
 Word Score, Residual                                                                                                                                                 &      21 & 0.0570744 & 8.04490438e-07    \\
 \hline
 SBT Score:                                                                                                                                                           &      22 & 0.0569778 & 1.40497951e-07    \\
 \hline
 Current Medications:  Confirmed with participant (choice=Benzodiazepines (diazepam, clonazepam, alprazolam))\_Unchecked                                               &      23 & 0.0568526 & 0   \\
 \hline
 Color-Word Score (CW), T-Score                                                                                                                                       &      24 & 0.0561718 & 1.36966936e-05    \\
 \hline
 Medical History:  Confirmed with participant (choice=Cancer or Hx Cancer)\_Unchecked                                                                                  &      25 & 0.056036  & 2.47380147e-09    \\
 \hline
 Medical History:  Confirmed with participant (choice=Cancer or Hx Cancer)\_Checked                                                                                    &      26 & 0.0560167 & 8.56659354e-10    \\
 \hline
 Medical History:  Confirmed with participant (choice=Heart Disease (CAD, CHF, PAD))\_Checked                                                                          &      27 & 0.0554194 & 5.05811537e-10    \\
 \hline
 Total Correct                                                                                                                                                        &      28 & 0.0552498 & 3.05141385e-05    \\
 \hline
 Medical History:  Confirmed with participant (choice=Diabetes)\_Checked                                                                                               &      29 & 0.0548924 & 3.0714916e-10    \\
 \hline
 Color-Word Score (CW), Age/Ed. Predicted                                                                                                                             &      30 & 0.0540521 & 4.95443076e-07    \\
 \hline
 Current Medications:  Confirmed with participant (choice=SNRI antidepressants (venlafaxine, duloxetine, desvenlafaxine, milnacipran))\_Checked                        &      31 & 0.0536455 & 2.90962588e-10    \\
 \hline
 Age (E-Consent):                                                                                                                                                     &      32 & 0.0534179 & 9.35483487e-09    \\
 \hline
 Final T-Score                                                                                                                                                        &      33 & 0.053231  & 2.39802603e-06    \\
 \hline
 Medical History:  Confirmed with participant (choice=Hypertension)\_Unchecked                                                                                         &      34 & 0.0532083 & 0   \\
 \hline
 Word Score, T-Score                                                                                                                                                  &      35 & 0.0532053 & 2.75284503e-06    \\
 \hline
 Do you take any of the following medications for pain treatment at least once per week? (choice=Duloxetine (Cymbalta) or amitriptyline (Elavil))\_Unchecked           &      36 & 0.0530448 & 0.00020982    \\
 \hline

\caption{Feature importance using \acrshort{mlp} using \gls{mi} scoring selecting features with scores $\sigma + 1\mu$.}
\label{tab:mrmr_mlp_linear}

\end{longtable}

\subsection{$\mu + 1\sigma$, \acrshort{sts}}

\begin{longtable}{|p{0.3\textwidth}|l|p{0.135\textwidth}|l|l|p{0.125\textwidth}|}
\hline
 Feature Name & $n$ & Score & Mean Absolute \acrshort{shap} \\ \hline

 On a scale of zero to ten, with zero being no pain and ten being the worst pain, please mark your average pain level during the past week, while at rest.                                        &       1 & 0.645585 & 2.52998313e-06    \\
 \hline
 On a scale of zero to ten, with zero being no pain and ten being the  worst pain , please mark your  average pain level during the past week, while you were active or moving.                   &       2 & 0.645342 & 4.85071774e-06    \\
 \hline
 Have you been experiencing pain in the past week?\_No                                                                                                                                             &       3 & 0.639949 & 0.00097941    \\
 \hline
 Have you been experiencing pain in the past week?\_Yes, and RELATED to my need for surgery                                                                                                        &       4 & 0.638486 & 0.00827323    \\
 \hline
 Have you been experiencing pain in the past week?\_Yes, but NOT RELATED to my need for surgery                                                                                                    &       5 & 0.634165 & 0.00756546    \\
 \hline
 On a scale of zero to ten, with zero being no pain and ten being the worst pain, what is your average pain level at rest TODAY?                                                                  &       6 & 0.619892 & 5.57362602e-06    \\
 \hline
 In the past three months, did you experience daily or near daily pain?\_No                                                                                                                        &       7 & 0.544496 & 0.00685028    \\
 \hline
 Have you ever been diagnosed with chronic pain?\_No                                                                                                                                               &       8 & 0.534078 & 0.0076836    \\
 \hline
 Have you ever been diagnosed with chronic pain?\_Yes                                                                                                                                              &       9 & 0.531654 & 0.00385837    \\
 \hline
 In the past three months, did you experience daily or near daily pain?\_Yes                                                                                                                       &      10 & 0.526172 & 0.00479166    \\
 \hline
 Pain intensity - Baseline Average:                                                                                                                                                               &      11 & 0.525021 & 2.64689293e-06    \\
 \hline
 Pain intensity - Baseline 1:                                                                                                                                                                     &      12 & 0.487709 & 6.1944453e-06    \\
 \hline
 Do you take any of the following medications for pain treatment at least once per week? (choice=None)\_Unchecked                                                                                  &      13 & 0.478695 & 0.00105593    \\
 \hline
 Pain intensity - Baseline 2:                                                                                                                                                                     &      14 & 0.475303 & 1.54633563e-06    \\
 \hline
 Do you take any of the following medications for pain treatment at least once per week? (choice=Other)\_Unchecked                                                                                 &      15 & 0.475231 & 0.00597048    \\
 \hline
 Do you take any of the following medications for pain treatment at least once per week? (choice=None)\_Checked                                                                                    &      16 & 0.461651 & 0.00330743    \\
 \hline
 Do you take any of the following medications for pain treatment at least once per week? (choice=Duloxetine (Cymbalta) or amitriptyline (Elavil))\_Unchecked                                       &      17 & 0.449812 & 0.00079825    \\
 \hline
 Have you used opioid medications for pain management in the past?\_I don't know                                                                                                                   &      18 & 0.44269  & 0.00037989    \\
 \hline
 Have you used opioid medications for pain management in the past?\_yes                                                                                                                            &      19 & 0.43422  & 0.00675297    \\
 \hline
 Do you take any of the following medications for pain treatment at least once per week? (choice=Other)\_Checked                                                                                   &      20 & 0.430055 & 0.00186429    \\
 \hline
 Have you used opioid medications for pain management in the past?\_no                                                                                                                             &      21 & 0.427981 & 0.00257769    \\
 \hline
 Do you take any of the following medications for pain treatment at least once per week? (choice=Gabapentin or Pregabalin)\_Unchecked                                                              &      22 & 0.417516 & 0.0130704    \\
 \hline
 Do you take any of the following medications for pain treatment at least once per week? (choice=Ibuprofen (Motrin or Advil), celecoxib (Celebrex, Naproxen or Aleve), or other NSAIDs)\_Unchecked &      23 & 0.416656 & 0.00558276    \\
 \hline
 Do you take any of the following medications for pain treatment at least once per week? (choice=Tramadol (Ultram))\_Unchecked                                                                     &      24 & 0.416248 & 0.00518861    \\
 \hline
 Do you take any of the following medications for pain treatment at least once per week? (choice=Acetaminophen (Tylenol))\_Unchecked                                                               &      25 & 0.415403 & 0.00048819    \\
 \hline
 Pain intensity - CPM Average:                                                                                                                                                                    &      26 & 0.413333 & 3.80922354e-06    \\
 \hline
 Do you take any of the following medications for pain treatment at least once per week? (choice=Duloxetine (Cymbalta) or amitriptyline (Elavil))\_Checked                                         &      27 & 0.40647  & 0.00236485    \\
 \hline
 Do you take any of the following medications for pain treatment at least once per week? (choice=Aspirin)\_Unchecked                                                                               &      28 & 0.397003 & 0.00082881    \\
 \hline
 Do you take any of the following medications for pain treatment at least once per week? (choice=Tramadol (Ultram))\_Checked                                                                       &      29 & 0.396227 & 0.0033335    \\
 \hline
 Pain intensity - CPM 1:                                                                                                                                                                          &      30 & 0.394309 &  8.07634382e-06    \\
 \hline
 Pain intensity - CPM 2:                                                                                                                                                                          &      31 & 0.393898 & 2.06799025e-06    \\
 \hline
 Do you take any of the following medications for pain treatment at least once per week? (choice=Acetaminophen (Tylenol))\_Checked                                                                 &      32 & 0.387619 & 0.00553789    \\
 \hline

\caption{Feature importance using \acrshort{mlp} using \gls{sts} scoring with PubMedBERT fine-tuned on the combined vocab selecting features with scores $\sigma + 1\mu$.}
\label{tab:mrmr_mlp_linear}

\end{longtable}

\subsection{$\mu + 1\sigma$, \acrshort{mi} \& \acrshort{sts}}

\begin{longtable}{|p{0.3\textwidth}|l|p{0.135\textwidth}|l|l|p{0.125\textwidth}|}
\hline
 Feature Name & $n$ & Score & Mean Absolute \acrshort{shap} \\ \hline
 Have you ever been diagnosed with chronic pain?\_No                                                                                                                                               &       1 & 0.734467 & 0.00860101         \\
 \hline
 Have you ever been diagnosed with chronic pain?\_Yes                                                                                                                                              &       2 & 0.712339 & 0.00457072         \\
 \hline
 Have you been experiencing pain in the past week?\_No                                                                                                                                             &       3 & 0.599768 & 7.51307387e-06 \\
 \hline
 Have you been experiencing pain in the past week?\_Yes, and RELATED to my need for surgery                                                                                                        &       4 & 0.598852 & 0.01351249         \\
 \hline
 Have you been experiencing pain in the past week?\_Yes, but NOT RELATED to my need for surgery                                                                                                    &       5 & 0.529267 & 6.29499593e-06 \\
 \hline
 In the past three months, did you experience daily or near daily pain?\_Yes                                                                                                                       &       6 & 0.525755 & 0.00436844         \\
 \hline
 Do you take any of the following medications for pain treatment at least once per week? (choice=Duloxetine (Cymbalta) or amitriptyline (Elavil))\_Unchecked                                       &       7 & 0.512279 & 3.79335761e-05 \\
 \hline
 In the past three months, did you experience daily or near daily pain?\_No                                                                                                                        &       8 & 0.504445 & 0.00791056         \\
 \hline
 Helplessness Score                                                                                                                                                                               &       9 & 0.493851 & 3.21395627e-06 \\
 \hline
 Pain intensity - Baseline 2:                                                                                                                                                                     &      10 & 0.479731 & 1.51890814e-06 \\
 \hline
 Pain intensity - Baseline Average:                                                                                                                                                               &      11 & 0.478815 & 4.47488069e-07 \\
 \hline
 Do you take any of the following medications for pain treatment at least once per week? (choice=Other)\_Unchecked                                                                                 &      12 & 0.472314 & 0.00774911         \\
 \hline
 Have you used opioid medications for pain management in the past?\_yes                                                                                                                            &      13 & 0.466008 & 0.00828783         \\
 \hline
 Pain intensity - CPM 1:                                                                                                                                                                          &      14 & 0.457419 & 1.58137085e-06 \\
 \hline
 Do you take any of the following medications for pain treatment at least once per week? (choice=None)\_Unchecked                                                                                  &      15 & 0.455803 & 0.00064399         \\
 \hline
 Pain intensity - CPM 2:                                                                                                                                                                          &      16 & 0.444577 & 3.07856456e-07 \\
 \hline
 Pain intensity - CPM Average:                                                                                                                                                                    &      17 & 0.439207 & 4.12713718e-06 \\
 \hline
 Do you take any of the following medications for pain treatment at least once per week? (choice=Tramadol (Ultram))\_Unchecked                                                                     &      18 & 0.438722 & 0.00609676         \\
 \hline
 Have you used opioid medications for pain management in the past?\_I don't know                                                                                                                   &      19 & 0.43665  & 0.00050837         \\
 \hline
 Have you used opioid medications for pain management in the past?\_no                                                                                                                             &      20 & 0.433569 & 0.00174238         \\
 \hline
 Do you take any of the following medications for pain treatment at least once per week? (choice=Duloxetine (Cymbalta) or amitriptyline (Elavil))\_Checked                                         &      21 & 0.41803  & 0.00318739         \\
 \hline
 Do you take any of the following medications for pain treatment at least once per week? (choice=Ibuprofen (Motrin or Advil), celecoxib (Celebrex, Naproxen or Aleve), or other NSAIDs)\_Unchecked &      22 & 0.415362 & 0.00432255         \\
 \hline
 Do you take any of the following medications for pain treatment at least once per week? (choice=Gabapentin or Pregabalin)\_Unchecked                                                              &      23 & 0.415204 & 0.01460327         \\
 \hline
 On a scale of zero to ten, with zero being no pain and ten being the worst pain, please mark your average pain level during the past week, while at rest.                                        &      24 & 0.404095 & 0                        \\
 \hline
 Do you take any of the following medications for pain treatment at least once per week? (choice=Aspirin)\_Unchecked                                                                               &      25 & 0.403687 & 0.00111563         \\
 \hline
 Current Medications:  Confirmed with participant (choice=Opioids (morphine, oxycodone, fentanyl, tramadol, methadone, buprenorphine))\_Unchecked                                                  &      26 & 0.398177 & 2.50144896e-06 \\
 \hline
 On a scale of zero to ten, with zero being no pain and ten being the worst pain, what is your average pain level at rest TODAY?                                                                  &      27 & 0.397393 & 1.43354317e-06 \\
 \hline

\caption{Feature importance using \acrshort{mlp} using linear scoring with PubMedBERT fine-tuned on the combined vocab selecting features with scores $\sigma + 1\mu$.}
\label{tab:mrmr_mlp_linear}

\end{longtable}

\section{Performance of All Feature Selection \& Classifier Combinations}

This section contains the \gls{auroc} and \gls{auprc} performance of all algorithms tested, separated by classifier. Maximum values for each classifier are bolded, as well as minimum absolute values for the test-train difference column. 

\subsection{Linear \gls{svm}}
\begin{longtable}{|l|l|l|l|l|l|l|l|l|l|}
\caption{All results for linear \gls{svm}.} \label{tab:LinearSVM} \\ \hline
\multicolumn{4}{|l|}{Selection Hyperparameters} & \multicolumn{3}{|l|}{AUROC} & \multicolumn{3}{|l|}{AUPRC} \\ \hline
Feature Selector & Scorer & Scoring Model & FT Dataset & Test & Train & $\Delta$ & Test & Train & $\Delta$ \\ \hline
\endfirsthead
\caption[]{All results for linear \gls{svm}.} \\ \hline
\multicolumn{4}{|l|}{Selection Hyperparameters} & \multicolumn{3}{|l|}{AUROC} & \multicolumn{3}{|l|}{AUPRC} \\ \hline
Feature Selector & Scorer & Scoring Model & FT Dataset & Test & Train & $\Delta$ & Test & Train & $\Delta$ \\ \hline
\endhead
\hline
\multicolumn{10}{r}{Continued on next page} \\ \hline
\endfoot
\hline
\endlastfoot
Identity & - & - & - & 0.534 & \textbf{0.858} & -0.324 & 0.033 & \textbf{0.463} & -0.430 \\ \hline
RFE & - & - & - & 0.521 & 0.826 & -0.305 & 0.032 & 0.438 & -0.406 \\ \hline
ReliefF & - & - & - & 0.581 & 0.785 & -0.203 & 0.037 & 0.174 & -0.138 \\ \hline
SFM-LinearSVM & - & - & - & 0.493 & 0.817 & -0.324 & 0.031 & 0.400 & -0.369 \\ \hline
SFM-XGBoost & - & - & - & 0.590 & 0.744 & -0.155 & 0.041 & 0.178 & -0.137 \\ \hline
Std. Dev. & MI & - & - & 0.689 & 0.712 & -0.024 & 0.049 & 0.116 & -0.067 \\ \hline
Std. Dev. & MI \& STS & Bio\_ClinicalBERT & ClinicalSTS & 0.561 & 0.782 & -0.221 & 0.039 & 0.358 & -0.319 \\ \hline
Std. Dev. & MI \& STS & Bio\_ClinicalBERT & clinical & 0.705 & 0.751 & -0.046 & 0.052 & 0.135 & -0.083 \\ \hline
Std. Dev. & MI \& STS & Bio\_ClinicalBERT & combined & 0.725 & 0.728 & -0.003 & 0.055 & 0.101 & -0.046 \\ \hline
Std. Dev. & MI \& STS & Bio\_ClinicalBERT & gen & 0.691 & 0.737 & -0.045 & 0.050 & 0.149 & -0.099 \\ \hline
Std. Dev. & MI \& STS & FastText & gen & 0.736 & 0.737 & \textbf{-0.002} & 0.101 & 0.108 & \textbf{-0.007} \\ \hline
Std. Dev. & MI \& STS & PubMedBERT & clinical & 0.686 & 0.763 & -0.077 & 0.048 & 0.204 & -0.155 \\ \hline
Std. Dev. & MI \& STS & PubMedBERT & combined & 0.628 & 0.770 & -0.142 & 0.045 & 0.321 & -0.277 \\ \hline
Std. Dev. & MI \& STS & PubMedBERT & gen & 0.628 & 0.770 & -0.142 & 0.041 & 0.284 & -0.243 \\ \hline
Std. Dev. & MI \& STS & Skipgram & gen & 0.661 & 0.759 & -0.098 & 0.044 & 0.166 & -0.122 \\ \hline
Std. Dev. & MI \& STS & all-MiniLM-L12-v2 & clinical & 0.544 & 0.771 & -0.227 & 0.036 & 0.338 & -0.302 \\ \hline
Std. Dev. & MI \& STS & all-MiniLM-L12-v2 & combined & 0.477 & 0.753 & -0.276 & 0.029 & 0.226 & -0.197 \\ \hline
Std. Dev. & MI \& STS & all-MiniLM-L12-v2 & gen & 0.645 & 0.784 & -0.140 & 0.044 & 0.164 & -0.120 \\ \hline
Std. Dev. & MI \& STS & bert-base-uncased & clinical & 0.618 & 0.692 & -0.073 & 0.047 & 0.105 & -0.058 \\ \hline
Std. Dev. & MI \& STS & bert-base-uncased & combined & 0.526 & 0.739 & -0.213 & 0.067 & 0.151 & -0.084 \\ \hline
Std. Dev. & MI \& STS & bert-base-uncased & gen & 0.634 & 0.733 & -0.099 & 0.056 & 0.121 & -0.065 \\ \hline
Std. Dev. & MI \& STS & biomed\_roberta\_base & clinical & 0.680 & 0.671 & 0.010 & 0.066 & 0.093 & -0.027 \\ \hline
Std. Dev. & MI \& STS & biomed\_roberta\_base & combined & 0.700 & 0.752 & -0.052 & 0.051 & 0.186 & -0.134 \\ \hline
Std. Dev. & MI \& STS & biomed\_roberta\_base & gen & 0.741 & 0.729 & 0.012 & 0.057 & 0.110 & -0.053 \\ \hline
Std. Dev. & STS & Bio\_ClinicalBERT & ClinicalSTS & 0.733 & 0.728 & 0.005 & 0.055 & 0.158 & -0.103 \\ \hline
Std. Dev. & STS & Bio\_ClinicalBERT & clinical & 0.642 & 0.707 & -0.065 & 0.042 & 0.161 & -0.119 \\ \hline
Std. Dev. & STS & Bio\_ClinicalBERT & combined & 0.708 & 0.758 & -0.050 & 0.051 & 0.207 & -0.156 \\ \hline
Std. Dev. & STS & Bio\_ClinicalBERT & gen & 0.713 & 0.753 & -0.040 & 0.052 & 0.194 & -0.142 \\ \hline
Std. Dev. & STS & FastText & gen & 0.559 & 0.806 & -0.247 & 0.035 & 0.249 & -0.213 \\ \hline
Std. Dev. & STS & PubMedBERT & clinical & 0.499 & 0.776 & -0.277 & 0.031 & 0.207 & -0.176 \\ \hline
Std. Dev. & STS & PubMedBERT & combined & 0.598 & 0.767 & -0.169 & 0.040 & 0.305 & -0.265 \\ \hline
Std. Dev. & STS & PubMedBERT & gen & 0.616 & 0.771 & -0.155 & 0.041 & 0.292 & -0.251 \\ \hline
Std. Dev. & STS & Skipgram & gen & 0.457 & 0.843 & -0.385 & 0.029 & 0.395 & -0.366 \\ \hline
Std. Dev. & STS & all-MiniLM-L12-v2 & clinical & 0.753 & 0.721 & 0.032 & 0.060 & 0.117 & -0.056 \\ \hline
Std. Dev. & STS & all-MiniLM-L12-v2 & combined & 0.508 & 0.785 & -0.276 & 0.032 & 0.297 & -0.265 \\ \hline
Std. Dev. & STS & all-MiniLM-L12-v2 & gen & 0.709 & 0.759 & -0.050 & 0.051 & 0.203 & -0.152 \\ \hline
Std. Dev. & STS & bert-base-uncased & clinical & 0.595 & 0.750 & -0.154 & 0.037 & 0.218 & -0.180 \\ \hline
Std. Dev. & STS & bert-base-uncased & combined & 0.614 & 0.769 & -0.155 & 0.039 & 0.268 & -0.229 \\ \hline
Std. Dev. & STS & bert-base-uncased & gen & 0.629 & 0.772 & -0.142 & 0.041 & 0.279 & -0.237 \\ \hline
Std. Dev. & STS & biomed\_roberta\_base & clinical & 0.559 & 0.805 & -0.246 & 0.035 & 0.237 & -0.202 \\ \hline
Std. Dev. & STS & biomed\_roberta\_base & combined & 0.579 & 0.786 & -0.207 & 0.037 & 0.282 & -0.244 \\ \hline
Std. Dev. & STS & biomed\_roberta\_base & gen & 0.606 & 0.775 & -0.169 & 0.039 & 0.243 & -0.203 \\ \hline
Top N & MI & - & - & 0.609 & 0.642 & -0.033 & 0.039 & 0.082 & -0.043 \\ \hline
Top N & MI \& STS & Bio\_ClinicalBERT & ClinicalSTS & 0.752 & 0.671 & 0.081 & 0.145 & 0.113 & 0.032 \\ \hline
Top N & MI \& STS & Bio\_ClinicalBERT & clinical & 0.672 & 0.745 & -0.073 & 0.047 & 0.117 & -0.071 \\ \hline
Top N & MI \& STS & Bio\_ClinicalBERT & combined & 0.664 & 0.753 & -0.089 & 0.046 & 0.152 & -0.105 \\ \hline
Top N & MI \& STS & Bio\_ClinicalBERT & gen & 0.683 & 0.747 & -0.064 & 0.050 & 0.133 & -0.083 \\ \hline
Top N & MI \& STS & FastText & gen & 0.565 & 0.681 & -0.116 & 0.035 & 0.099 & -0.064 \\ \hline
Top N & MI \& STS & PubMedBERT & clinical & 0.713 & 0.733 & -0.020 & 0.052 & 0.159 & -0.107 \\ \hline
Top N & MI \& STS & PubMedBERT & combined & 0.769 & 0.721 & 0.047 & 0.070 & 0.140 & -0.070 \\ \hline
Top N & MI \& STS & PubMedBERT & gen & 0.563 & 0.776 & -0.212 & 0.037 & 0.238 & -0.201 \\ \hline
Top N & MI \& STS & Skipgram & gen & \textbf{0.782} & 0.594 & \textbf{0.188} & 0.068 & 0.077 & -0.009 \\ \hline
Top N & MI \& STS & all-MiniLM-L12-v2 & clinical & 0.696 & 0.721 & -0.025 & 0.049 & 0.131 & -0.082 \\ \hline
Top N & MI \& STS & all-MiniLM-L12-v2 & combined & 0.601 & 0.733 & -0.132 & 0.038 & 0.180 & -0.143 \\ \hline
Top N & MI \& STS & all-MiniLM-L12-v2 & gen & 0.661 & 0.760 & -0.098 & 0.045 & 0.159 & -0.114 \\ \hline
Top N & MI \& STS & bert-base-uncased & clinical & 0.594 & 0.682 & -0.088 & 0.053 & 0.091 & -0.038 \\ \hline
Top N & MI \& STS & bert-base-uncased & combined & 0.594 & 0.698 & -0.104 & 0.053 & 0.131 & -0.077 \\ \hline
Top N & MI \& STS & bert-base-uncased & gen & 0.574 & 0.734 & -0.159 & 0.037 & 0.139 & -0.102 \\ \hline
Top N & MI \& STS & biomed\_roberta\_base & clinical & 0.639 & 0.648 & -0.009 & 0.051 & 0.077 & -0.026 \\ \hline
Top N & MI \& STS & biomed\_roberta\_base & combined & 0.730 & 0.724 & 0.006 & 0.066 & 0.165 & -0.099 \\ \hline
Top N & MI \& STS & biomed\_roberta\_base & gen & 0.741 & 0.729 & 0.012 & 0.057 & 0.110 & -0.053 \\ \hline
Top N & STS & Bio\_ClinicalBERT & ClinicalSTS & 0.713 & 0.737 & -0.023 & 0.050 & 0.150 & -0.100 \\ \hline
Top N & STS & Bio\_ClinicalBERT & clinical & 0.680 & 0.728 & -0.048 & 0.046 & 0.162 & -0.115 \\ \hline
Top N & STS & Bio\_ClinicalBERT & combined & 0.675 & 0.713 & -0.038 & 0.046 & 0.151 & -0.106 \\ \hline
Top N & STS & Bio\_ClinicalBERT & gen & 0.649 & 0.727 & -0.078 & 0.043 & 0.145 & -0.102 \\ \hline
Top N & STS & FastText & gen & 0.642 & 0.723 & -0.081 & 0.042 & 0.146 & -0.104 \\ \hline
Top N & STS & PubMedBERT & clinical & 0.769 & 0.724 & 0.045 & 0.065 & 0.133 & -0.069 \\ \hline
Top N & STS & PubMedBERT & combined & 0.758 & 0.753 & 0.004 & 0.067 & 0.171 & -0.103 \\ \hline
Top N & STS & PubMedBERT & gen & 0.646 & 0.749 & -0.103 & 0.044 & 0.161 & -0.117 \\ \hline
Top N & STS & Skipgram & gen & 0.725 & 0.721 & 0.004 & 0.054 & 0.130 & -0.076 \\ \hline
Top N & STS & all-MiniLM-L12-v2 & clinical & 0.711 & 0.746 & -0.035 & 0.055 & 0.154 & -0.099 \\ \hline
Top N & STS & all-MiniLM-L12-v2 & combined & 0.617 & 0.735 & -0.118 & 0.040 & 0.135 & -0.096 \\ \hline
Top N & STS & all-MiniLM-L12-v2 & gen & 0.755 & 0.726 & 0.029 & 0.059 & 0.112 & -0.053 \\ \hline
Top N & STS & bert-base-uncased & clinical & 0.646 & 0.726 & -0.080 & 0.043 & 0.183 & -0.141 \\ \hline
Top N & STS & bert-base-uncased & combined & 0.517 & 0.754 & -0.238 & 0.030 & 0.189 & -0.159 \\ \hline
Top N & STS & bert-base-uncased & gen & 0.653 & 0.740 & -0.087 & 0.043 & 0.182 & -0.139 \\ \hline
Top N & STS & biomed\_roberta\_base & clinical & 0.533 & 0.737 & -0.204 & 0.038 & 0.240 & -0.202 \\ \hline
Top N & STS & biomed\_roberta\_base & combined & 0.723 & 0.714 & 0.009 & 0.059 & 0.116 & -0.057 \\ \hline
Top N & STS & biomed\_roberta\_base & gen & 0.748 & 0.720 & 0.028 & 0.068 & 0.168 & -0.100 \\ \hline
mRMR & MI & - & - & 0.391 & 0.671 & -0.280 & 0.025 & 0.092 & -0.067 \\ \hline
mRMR & MI \& STS & Bio\_ClinicalBERT & ClinicalSTS & 0.687 & 0.750 & -0.062 & 0.048 & 0.126 & -0.079 \\ \hline
mRMR & MI \& STS & Bio\_ClinicalBERT & clinical & 0.602 & 0.749 & -0.147 & 0.037 & 0.161 & -0.124 \\ \hline
mRMR & MI \& STS & Bio\_ClinicalBERT & combined & 0.576 & 0.746 & -0.170 & 0.036 & 0.151 & -0.115 \\ \hline
mRMR & MI \& STS & Bio\_ClinicalBERT & gen & 0.610 & 0.732 & -0.122 & 0.039 & 0.146 & -0.107 \\ \hline
mRMR & MI \& STS & FastText & gen & 0.234 & 0.736 & -0.502 & 0.021 & 0.153 & -0.132 \\ \hline
mRMR & MI \& STS & PubMedBERT & clinical & 0.614 & 0.757 & -0.143 & 0.041 & 0.168 & -0.127 \\ \hline
mRMR & MI \& STS & PubMedBERT & combined & 0.663 & 0.765 & -0.103 & 0.053 & 0.256 & -0.203 \\ \hline
mRMR & MI \& STS & PubMedBERT & gen & 0.679 & 0.799 & -0.120 & 0.063 & 0.301 & -0.239 \\ \hline
mRMR & MI \& STS & Skipgram & gen & 0.410 & 0.599 & -0.188 & 0.028 & 0.099 & -0.071 \\ \hline
mRMR & MI \& STS & all-MiniLM-L12-v2 & clinical & 0.603 & 0.673 & -0.070 & 0.039 & 0.103 & -0.064 \\ \hline
mRMR & MI \& STS & all-MiniLM-L12-v2 & combined & 0.620 & 0.775 & -0.155 & 0.040 & 0.150 & -0.110 \\ \hline
mRMR & MI \& STS & all-MiniLM-L12-v2 & gen & 0.682 & 0.779 & -0.097 & 0.048 & 0.156 & -0.109 \\ \hline
mRMR & MI \& STS & bert-base-uncased & clinical & 0.584 & 0.662 & -0.078 & 0.038 & 0.095 & -0.057 \\ \hline
mRMR & MI \& STS & bert-base-uncased & combined & 0.683 & 0.704 & -0.021 & 0.050 & 0.100 & -0.050 \\ \hline
mRMR & MI \& STS & bert-base-uncased & gen & 0.530 & 0.685 & -0.155 & 0.033 & 0.117 & -0.084 \\ \hline
mRMR & MI \& STS & biomed\_roberta\_base & clinical & 0.519 & 0.638 & -0.119 & 0.041 & 0.069 & -0.027 \\ \hline
mRMR & MI \& STS & biomed\_roberta\_base & combined & 0.686 & 0.665 & 0.020 & \textbf{0.195} & 0.103 & \textbf{0.092} \\ \hline
mRMR & MI \& STS & biomed\_roberta\_base & gen & 0.690 & 0.750 & -0.060 & 0.048 & 0.126 & -0.078 \\ \hline
mRMR & STS & Bio\_ClinicalBERT & ClinicalSTS & 0.708 & 0.702 & 0.006 & 0.052 & 0.118 & -0.066 \\ \hline
mRMR & STS & Bio\_ClinicalBERT & clinical & 0.640 & 0.757 & -0.117 & 0.042 & 0.156 & -0.114 \\ \hline
mRMR & STS & Bio\_ClinicalBERT & combined & 0.457 & 0.757 & -0.300 & 0.029 & 0.235 & -0.206 \\ \hline
mRMR & STS & Bio\_ClinicalBERT & gen & 0.506 & 0.739 & -0.233 & 0.032 & 0.199 & -0.167 \\ \hline
mRMR & STS & FastText & gen & 0.565 & 0.779 & -0.214 & 0.035 & 0.221 & -0.186 \\ \hline
mRMR & STS & PubMedBERT & clinical & 0.544 & 0.769 & -0.225 & 0.033 & 0.198 & -0.165 \\ \hline
mRMR & STS & PubMedBERT & combined & 0.620 & 0.724 & -0.104 & 0.041 & 0.191 & -0.150 \\ \hline
mRMR & STS & PubMedBERT & gen & 0.554 & 0.741 & -0.187 & 0.041 & 0.194 & -0.154 \\ \hline
mRMR & STS & Skipgram & gen & 0.581 & 0.803 & -0.222 & 0.037 & 0.360 & -0.324 \\ \hline
mRMR & STS & all-MiniLM-L12-v2 & clinical & 0.594 & 0.737 & -0.143 & 0.042 & 0.146 & -0.104 \\ \hline
mRMR & STS & all-MiniLM-L12-v2 & combined & 0.526 & 0.692 & -0.166 & 0.032 & 0.103 & -0.071 \\ \hline
mRMR & STS & all-MiniLM-L12-v2 & gen & 0.569 & 0.743 & -0.174 & 0.039 & 0.136 & -0.097 \\ \hline
mRMR & STS & bert-base-uncased & clinical & 0.628 & 0.747 & -0.119 & 0.041 & 0.113 & -0.072 \\ \hline
mRMR & STS & bert-base-uncased & combined & 0.493 & 0.732 & -0.239 & 0.031 & 0.135 & -0.104 \\ \hline
mRMR & STS & bert-base-uncased & gen & 0.579 & 0.721 & -0.143 & 0.038 & 0.085 & -0.048 \\ \hline
mRMR & STS & biomed\_roberta\_base & clinical & 0.523 & 0.762 & -0.239 & 0.036 & 0.140 & -0.104 \\ \hline
mRMR & STS & biomed\_roberta\_base & combined & 0.722 & 0.693 & 0.029 & 0.054 & 0.105 & -0.051 \\ \hline
mRMR & STS & biomed\_roberta\_base & gen & 0.587 & 0.703 & -0.116 & 0.036 & 0.109 & -0.072 \\ \hline
\end{longtable}

\subsection{XGBoost}
\begin{longtable}{|l|l|l|l|l|l|l|l|l|l|}
\caption{All results for XGBoost.} \label{tab:XGBoost} \\ \hline
\multicolumn{4}{|l|}{Selection Hyperparameters} & \multicolumn{3}{|l|}{AUROC} & \multicolumn{3}{|l|}{AUPRC} \\ \hline
Feature Selector & Scorer & Scoring Model & FT Dataset & Test & Train & $\Delta$ & Test & Train & $\Delta$ \\ \hline
\endfirsthead
\caption[]{All results for XGBoost.} \\ \hline
\multicolumn{4}{|l|}{Selection Hyperparameters} & \multicolumn{3}{|l|}{AUROC} & \multicolumn{3}{|l|}{AUPRC} \\ \hline
Feature Selector & Scorer & Scoring Model & FT Dataset & Test & Train & $\Delta$ & Test & Train & $\Delta$ \\ \hline
\endhead
\hline
\multicolumn{10}{r}{Continued on next page} \\ \hline
\endfoot
\hline
\endlastfoot
Identity & - & - & - & 0.488 & 0.784 & -0.296 & 0.024 & 0.338 & -0.314 \\ \hline
RFE & - & - & - & 0.417 & 0.749 & -0.332 & 0.024 & 0.129 & -0.105 \\ \hline
ReliefF & - & - & - & 0.483 & 0.727 & -0.244 & 0.024 & \textbf{0.479} & -0.455 \\ \hline
SFM-LinearSVM & - & - & - & 0.421 & 0.728 & -0.306 & 0.024 & 0.118 & -0.094 \\ \hline
SFM-XGBoost & - & - & - & 0.742 & \textbf{0.860} & -0.118 & 0.064 & 0.238 & -0.175 \\ \hline
Std. Dev. & MI & - & - & 0.450 & 0.761 & -0.311 & 0.024 & 0.305 & -0.281 \\ \hline
Std. Dev. & MI \& STS & Bio\_ClinicalBERT & ClinicalSTS & 0.455 & 0.637 & -0.182 & 0.024 & 0.110 & -0.086 \\ \hline
Std. Dev. & MI \& STS & Bio\_ClinicalBERT & clinical & 0.500 & 0.589 & -0.089 & 0.024 & 0.158 & -0.134 \\ \hline
Std. Dev. & MI \& STS & Bio\_ClinicalBERT & combined & 0.500 & 0.589 & -0.089 & 0.024 & 0.158 & -0.134 \\ \hline
Std. Dev. & MI \& STS & Bio\_ClinicalBERT & gen & 0.500 & 0.589 & -0.089 & 0.024 & 0.158 & -0.134 \\ \hline
Std. Dev. & MI \& STS & FastText & gen & 0.450 & 0.761 & -0.311 & 0.024 & 0.305 & -0.281 \\ \hline
Std. Dev. & MI \& STS & PubMedBERT & clinical & 0.617 & 0.750 & -0.133 & 0.042 & 0.219 & -0.177 \\ \hline
Std. Dev. & MI \& STS & PubMedBERT & combined & 0.467 & 0.768 & -0.302 & 0.024 & 0.429 & -0.405 \\ \hline
Std. Dev. & MI \& STS & PubMedBERT & gen & 0.479 & 0.763 & -0.284 & 0.024 & 0.332 & -0.308 \\ \hline
Std. Dev. & MI \& STS & Skipgram & gen & 0.530 & 0.774 & -0.244 & 0.026 & 0.117 & -0.091 \\ \hline
Std. Dev. & MI \& STS & all-MiniLM-L12-v2 & clinical & 0.483 & 0.628 & -0.144 & 0.024 & 0.149 & -0.125 \\ \hline
Std. Dev. & MI \& STS & all-MiniLM-L12-v2 & combined & 0.625 & 0.688 & -0.062 & 0.046 & 0.174 & -0.127 \\ \hline
Std. Dev. & MI \& STS & all-MiniLM-L12-v2 & gen & 0.475 & 0.609 & -0.134 & 0.024 & 0.161 & -0.137 \\ \hline
Std. Dev. & MI \& STS & bert-base-uncased & clinical & 0.492 & 0.562 & -0.070 & 0.024 & 0.084 & -0.060 \\ \hline
Std. Dev. & MI \& STS & bert-base-uncased & combined & 0.467 & 0.738 & -0.271 & 0.024 & 0.272 & -0.248 \\ \hline
Std. Dev. & MI \& STS & bert-base-uncased & gen & 0.463 & 0.670 & -0.207 & 0.024 & 0.114 & -0.090 \\ \hline
Std. Dev. & MI \& STS & biomed\_roberta\_base & clinical & 0.479 & 0.636 & -0.157 & 0.024 & 0.305 & -0.281 \\ \hline
Std. Dev. & MI \& STS & biomed\_roberta\_base & combined & 0.442 & 0.736 & -0.294 & 0.024 & 0.251 & -0.227 \\ \hline
Std. Dev. & MI \& STS & biomed\_roberta\_base & gen & 0.563 & 0.637 & -0.074 & 0.029 & 0.072 & -0.043 \\ \hline
Std. Dev. & STS & Bio\_ClinicalBERT & ClinicalSTS & 0.434 & 0.701 & -0.267 & 0.024 & 0.119 & -0.095 \\ \hline
Std. Dev. & STS & Bio\_ClinicalBERT & clinical & 0.455 & 0.629 & -0.175 & 0.024 & 0.094 & -0.070 \\ \hline
Std. Dev. & STS & Bio\_ClinicalBERT & combined & 0.450 & 0.718 & -0.268 & 0.024 & 0.156 & -0.132 \\ \hline
Std. Dev. & STS & Bio\_ClinicalBERT & gen & 0.434 & 0.701 & -0.267 & 0.024 & 0.119 & -0.095 \\ \hline
Std. Dev. & STS & FastText & gen & 0.450 & 0.838 & -0.388 & 0.024 & 0.303 & -0.279 \\ \hline
Std. Dev. & STS & PubMedBERT & clinical & 0.426 & 0.639 & -0.213 & 0.024 & 0.078 & -0.054 \\ \hline
Std. Dev. & STS & PubMedBERT & combined & 0.426 & 0.639 & -0.213 & 0.024 & 0.078 & -0.054 \\ \hline
Std. Dev. & STS & PubMedBERT & gen & 0.426 & 0.639 & -0.213 & 0.024 & 0.078 & -0.054 \\ \hline
Std. Dev. & STS & Skipgram & gen & 0.471 & 0.744 & -0.273 & 0.024 & 0.346 & -0.322 \\ \hline
Std. Dev. & STS & all-MiniLM-L12-v2 & clinical & 0.479 & 0.596 & -0.116 & 0.024 & 0.086 & -0.062 \\ \hline
Std. Dev. & STS & all-MiniLM-L12-v2 & combined & 0.455 & 0.637 & -0.182 & 0.024 & 0.110 & -0.086 \\ \hline
Std. Dev. & STS & all-MiniLM-L12-v2 & gen & 0.434 & 0.701 & -0.267 & 0.024 & 0.119 & -0.095 \\ \hline
Std. Dev. & STS & bert-base-uncased & clinical & 0.455 & 0.637 & -0.182 & 0.024 & 0.110 & -0.086 \\ \hline
Std. Dev. & STS & bert-base-uncased & combined & 0.455 & 0.637 & -0.182 & 0.024 & 0.110 & -0.086 \\ \hline
Std. Dev. & STS & bert-base-uncased & gen & 0.426 & 0.639 & -0.213 & 0.024 & 0.078 & -0.054 \\ \hline
Std. Dev. & STS & biomed\_roberta\_base & clinical & 0.471 & 0.628 & -0.157 & 0.024 & 0.149 & -0.125 \\ \hline
Std. Dev. & STS & biomed\_roberta\_base & combined & 0.617 & 0.748 & -0.131 & 0.042 & 0.207 & -0.166 \\ \hline
Std. Dev. & STS & biomed\_roberta\_base & gen & 0.430 & 0.639 & -0.209 & 0.024 & 0.078 & -0.054 \\ \hline
Top N & MI & - & - & 0.471 & 0.633 & -0.162 & 0.024 & 0.214 & -0.190 \\ \hline
Top N & MI \& STS & Bio\_ClinicalBERT & ClinicalSTS & 0.471 & 0.587 & -0.116 & 0.024 & 0.072 & -0.048 \\ \hline
Top N & MI \& STS & Bio\_ClinicalBERT & clinical & 0.500 & 0.589 & -0.089 & 0.024 & 0.158 & -0.134 \\ \hline
Top N & MI \& STS & Bio\_ClinicalBERT & combined & 0.500 & 0.589 & -0.089 & 0.024 & 0.158 & -0.134 \\ \hline
Top N & MI \& STS & Bio\_ClinicalBERT & gen & 0.500 & 0.589 & -0.089 & 0.024 & 0.158 & -0.134 \\ \hline
Top N & MI \& STS & FastText & gen & 0.450 & 0.743 & -0.293 & 0.024 & 0.184 & -0.160 \\ \hline
Top N & MI \& STS & PubMedBERT & clinical & 0.467 & 0.738 & -0.271 & 0.024 & 0.272 & -0.248 \\ \hline
Top N & MI \& STS & PubMedBERT & combined & 0.463 & 0.715 & -0.252 & 0.024 & 0.231 & -0.207 \\ \hline
Top N & MI \& STS & PubMedBERT & gen & 0.475 & 0.681 & -0.206 & 0.024 & 0.145 & -0.121 \\ \hline
Top N & MI \& STS & Skipgram & gen & 0.467 & 0.669 & -0.202 & 0.024 & 0.174 & -0.150 \\ \hline
Top N & MI \& STS & all-MiniLM-L12-v2 & clinical & 0.467 & 0.579 & -0.112 & 0.024 & 0.085 & -0.061 \\ \hline
Top N & MI \& STS & all-MiniLM-L12-v2 & combined & 0.467 & 0.579 & -0.112 & 0.024 & 0.085 & -0.061 \\ \hline
Top N & MI \& STS & all-MiniLM-L12-v2 & gen & 0.479 & 0.633 & -0.154 & 0.024 & 0.214 & -0.190 \\ \hline
Top N & MI \& STS & bert-base-uncased & clinical & 0.500 & 0.500 & \textbf{0.000} & 0.024 & 0.045 & \textbf{-0.020} \\ \hline
Top N & MI \& STS & bert-base-uncased & combined & 0.471 & 0.703 & -0.232 & 0.024 & 0.162 & -0.138 \\ \hline
Top N & MI \& STS & bert-base-uncased & gen & 0.463 & 0.674 & -0.211 & 0.024 & 0.123 & -0.099 \\ \hline
Top N & MI \& STS & biomed\_roberta\_base & clinical & 0.479 & 0.636 & -0.157 & 0.024 & 0.305 & -0.281 \\ \hline
Top N & MI \& STS & biomed\_roberta\_base & combined & 0.475 & 0.651 & -0.175 & 0.024 & 0.179 & -0.155 \\ \hline
Top N & MI \& STS & biomed\_roberta\_base & gen & 0.563 & 0.637 & -0.074 & 0.029 & 0.072 & -0.043 \\ \hline
Top N & STS & Bio\_ClinicalBERT & ClinicalSTS & 0.563 & 0.637 & -0.074 & 0.029 & 0.072 & -0.043 \\ \hline
Top N & STS & Bio\_ClinicalBERT & clinical & 0.467 & 0.579 & -0.112 & 0.024 & 0.085 & -0.061 \\ \hline
Top N & STS & Bio\_ClinicalBERT & combined & 0.479 & 0.585 & -0.105 & 0.024 & 0.109 & -0.085 \\ \hline
Top N & STS & Bio\_ClinicalBERT & gen & 0.467 & 0.579 & -0.112 & 0.024 & 0.085 & -0.061 \\ \hline
Top N & STS & FastText & gen & 0.496 & 0.639 & -0.143 & 0.024 & 0.116 & -0.092 \\ \hline
Top N & STS & PubMedBERT & clinical & 0.479 & 0.596 & -0.116 & 0.024 & 0.086 & -0.062 \\ \hline
Top N & STS & PubMedBERT & combined & 0.479 & 0.596 & -0.116 & 0.024 & 0.086 & -0.062 \\ \hline
Top N & STS & PubMedBERT & gen & 0.471 & 0.676 & -0.205 & 0.024 & 0.129 & -0.104 \\ \hline
Top N & STS & Skipgram & gen & 0.496 & 0.639 & -0.143 & 0.024 & 0.116 & -0.092 \\ \hline
Top N & STS & all-MiniLM-L12-v2 & clinical & 0.563 & 0.637 & -0.074 & 0.029 & 0.072 & -0.043 \\ \hline
Top N & STS & all-MiniLM-L12-v2 & combined & 0.479 & 0.596 & -0.116 & 0.024 & 0.086 & -0.062 \\ \hline
Top N & STS & all-MiniLM-L12-v2 & gen & 0.563 & 0.637 & -0.074 & 0.029 & 0.072 & -0.043 \\ \hline
Top N & STS & bert-base-uncased & clinical & 0.467 & 0.579 & -0.112 & 0.024 & 0.085 & -0.061 \\ \hline
Top N & STS & bert-base-uncased & combined & 0.430 & 0.639 & -0.209 & 0.024 & 0.078 & -0.054 \\ \hline
Top N & STS & bert-base-uncased & gen & 0.426 & 0.694 & -0.268 & 0.024 & 0.108 & -0.084 \\ \hline
Top N & STS & biomed\_roberta\_base & clinical & 0.450 & 0.619 & -0.169 & 0.024 & 0.107 & -0.083 \\ \hline
Top N & STS & biomed\_roberta\_base & combined & 0.479 & 0.596 & -0.116 & 0.024 & 0.086 & -0.062 \\ \hline
Top N & STS & biomed\_roberta\_base & gen & 0.479 & 0.596 & -0.116 & 0.024 & 0.086 & -0.062 \\ \hline
mRMR & MI & - & - & 0.492 & 0.654 & -0.162 & 0.024 & 0.216 & -0.192 \\ \hline
mRMR & MI \& STS & Bio\_ClinicalBERT & ClinicalSTS & 0.563 & 0.637 & -0.074 & 0.029 & 0.072 & -0.043 \\ \hline
mRMR & MI \& STS & Bio\_ClinicalBERT & clinical & 0.388 & 0.709 & -0.320 & 0.024 & 0.092 & -0.068 \\ \hline
mRMR & MI \& STS & Bio\_ClinicalBERT & combined & 0.388 & 0.709 & -0.320 & 0.024 & 0.092 & -0.068 \\ \hline
mRMR & MI \& STS & Bio\_ClinicalBERT & gen & 0.421 & 0.719 & -0.298 & 0.024 & 0.099 & -0.075 \\ \hline
mRMR & MI \& STS & FastText & gen & 0.467 & 0.735 & -0.268 & 0.024 & 0.242 & -0.218 \\ \hline
mRMR & MI \& STS & PubMedBERT & clinical & 0.488 & 0.628 & -0.140 & 0.024 & 0.149 & -0.125 \\ \hline
mRMR & MI \& STS & PubMedBERT & combined & 0.471 & 0.663 & -0.192 & 0.024 & 0.140 & -0.116 \\ \hline
mRMR & MI \& STS & PubMedBERT & gen & 0.475 & 0.667 & -0.192 & 0.024 & 0.161 & -0.136 \\ \hline
mRMR & MI \& STS & Skipgram & gen & 0.483 & 0.631 & -0.148 & 0.024 & 0.181 & -0.157 \\ \hline
mRMR & MI \& STS & all-MiniLM-L12-v2 & clinical & 0.563 & 0.637 & -0.074 & 0.029 & 0.072 & -0.043 \\ \hline
mRMR & MI \& STS & all-MiniLM-L12-v2 & combined & 0.526 & 0.717 & -0.191 & 0.026 & 0.088 & -0.062 \\ \hline
mRMR & MI \& STS & all-MiniLM-L12-v2 & gen & 0.563 & 0.637 & -0.074 & 0.029 & 0.072 & -0.043 \\ \hline
mRMR & MI \& STS & bert-base-uncased & clinical & 0.500 & 0.500 & \textbf{0.000} & 0.024 & 0.045 & \textbf{-0.020} \\ \hline
mRMR & MI \& STS & bert-base-uncased & combined & 0.459 & 0.562 & -0.104 & 0.024 & 0.060 & -0.036 \\ \hline
mRMR & MI \& STS & bert-base-uncased & gen & 0.471 & 0.575 & -0.104 & 0.024 & 0.075 & -0.051 \\ \hline
mRMR & MI \& STS & biomed\_roberta\_base & clinical & 0.488 & 0.534 & -0.046 & 0.024 & 0.055 & -0.030 \\ \hline
mRMR & MI \& STS & biomed\_roberta\_base & combined & 0.479 & 0.589 & -0.109 & 0.024 & 0.158 & -0.134 \\ \hline
mRMR & MI \& STS & biomed\_roberta\_base & gen & 0.459 & 0.667 & -0.209 & 0.024 & 0.110 & -0.086 \\ \hline
mRMR & STS & Bio\_ClinicalBERT & ClinicalSTS & 0.613 & 0.707 & -0.094 & 0.040 & 0.181 & -0.141 \\ \hline
mRMR & STS & Bio\_ClinicalBERT & clinical & 0.563 & 0.637 & -0.074 & 0.029 & 0.072 & -0.043 \\ \hline
mRMR & STS & Bio\_ClinicalBERT & combined & 0.634 & 0.690 & -0.056 & 0.053 & 0.186 & -0.133 \\ \hline
mRMR & STS & Bio\_ClinicalBERT & gen & 0.479 & 0.586 & -0.106 & 0.024 & 0.117 & -0.093 \\ \hline
mRMR & STS & FastText & gen & 0.426 & 0.807 & -0.382 & 0.024 & 0.181 & -0.157 \\ \hline
mRMR & STS & PubMedBERT & clinical & 0.459 & 0.728 & -0.269 & 0.024 & 0.194 & -0.170 \\ \hline
mRMR & STS & PubMedBERT & combined & 0.500 & 0.589 & -0.089 & 0.024 & 0.158 & -0.134 \\ \hline
mRMR & STS & PubMedBERT & gen & 0.455 & 0.678 & -0.223 & 0.024 & 0.135 & -0.110 \\ \hline
mRMR & STS & Skipgram & gen & 0.459 & 0.764 & -0.306 & 0.024 & 0.348 & -0.323 \\ \hline
mRMR & STS & all-MiniLM-L12-v2 & clinical & 0.522 & 0.738 & -0.216 & 0.025 & 0.094 & -0.068 \\ \hline
mRMR & STS & all-MiniLM-L12-v2 & combined & 0.522 & 0.738 & -0.216 & 0.025 & 0.094 & -0.068 \\ \hline
mRMR & STS & all-MiniLM-L12-v2 & gen & 0.430 & 0.735 & -0.306 & 0.024 & 0.113 & -0.089 \\ \hline
mRMR & STS & bert-base-uncased & clinical & 0.500 & 0.590 & -0.090 & 0.024 & 0.182 & -0.158 \\ \hline
mRMR & STS & bert-base-uncased & combined & 0.483 & 0.694 & -0.210 & 0.024 & 0.220 & -0.196 \\ \hline
mRMR & STS & bert-base-uncased & gen & \textbf{0.747} & 0.726 & \textbf{0.021} & \textbf{0.066} & 0.104 & -0.038 \\ \hline
mRMR & STS & biomed\_roberta\_base & clinical & 0.479 & 0.692 & -0.212 & 0.024 & 0.202 & -0.178 \\ \hline
mRMR & STS & biomed\_roberta\_base & combined & 0.434 & 0.786 & -0.352 & 0.024 & 0.219 & -0.195 \\ \hline
mRMR & STS & biomed\_roberta\_base & gen & 0.563 & 0.637 & -0.074 & 0.029 & 0.072 & -0.043 \\ \hline
\end{longtable}

\subsection{\gls{mlp}}
\begin{longtable}{|l|l|l|l|l|l|l|l|l|l|}
\caption{All results for MLP.} \label{tab:MLP} \\ \hline
\multicolumn{4}{|l|}{Selection Hyperparameters} & \multicolumn{3}{|l|}{AUROC} & \multicolumn{3}{|l|}{AUPRC} \\ \hline
Feature Selector & Scorer & Scoring Model & FT Dataset & Test & Train & $\Delta$ & Test & Train & $\Delta$ \\ \hline
\endfirsthead
\caption[]{All results for MLP.} \\ \hline
\multicolumn{4}{|l|}{Selection Hyperparameters} & \multicolumn{3}{|l|}{AUROC} & \multicolumn{3}{|l|}{AUPRC} \\ \hline
Feature Selector & Scorer & Scoring Model & FT Dataset & Test & Train & $\Delta$ & Test & Train & $\Delta$ \\ \hline
\endhead
\hline
\multicolumn{10}{r}{Continued on next page} \\ \hline
\endfoot
\hline
\endlastfoot
Identity & - & - & - & 0.576 & 0.881 & -0.305 & 0.036 & 0.541 & -0.505 \\ \hline
RFE & - & - & - & 0.534 & 0.834 & -0.299 & 0.034 & 0.452 & -0.418 \\ \hline
ReliefF & - & - & - & 0.545 & 0.861 & -0.315 & 0.034 & 0.348 & -0.314 \\ \hline
SFM-LinearSVM & - & - & - & 0.658 & 0.791 & -0.132 & 0.045 & 0.286 & -0.242 \\ \hline
SFM-XGBoost & - & - & - & 0.430 & 0.679 & -0.249 & 0.028 & 0.151 & -0.123 \\ \hline
Std. Dev. & MI & - & - & 0.259 & 0.494 & -0.235 & 0.021 & 0.044 & -0.022 \\ \hline
Std. Dev. & MI \& STS & Bio\_ClinicalBERT & ClinicalSTS & 0.638 & 0.786 & -0.148 & 0.042 & 0.336 & -0.294 \\ \hline
Std. Dev. & MI \& STS & Bio\_ClinicalBERT & clinical & 0.689 & 0.765 & -0.077 & 0.049 & 0.143 & -0.094 \\ \hline
Std. Dev. & MI \& STS & Bio\_ClinicalBERT & combined & 0.705 & 0.776 & -0.071 & 0.052 & 0.174 & -0.122 \\ \hline
Std. Dev. & MI \& STS & Bio\_ClinicalBERT & gen & 0.650 & 0.770 & -0.120 & 0.044 & 0.218 & -0.174 \\ \hline
Std. Dev. & MI \& STS & FastText & gen & 0.738 & 0.758 & -0.019 & \textbf{0.205} & 0.123 & \textbf{0.082} \\ \hline
Std. Dev. & MI \& STS & PubMedBERT & clinical & 0.631 & 0.786 & -0.155 & 0.041 & 0.278 & -0.236 \\ \hline
Std. Dev. & MI \& STS & PubMedBERT & combined & 0.689 & 0.774 & -0.085 & 0.049 & 0.316 & -0.267 \\ \hline
Std. Dev. & MI \& STS & PubMedBERT & gen & 0.645 & 0.788 & -0.144 & 0.044 & 0.319 & -0.275 \\ \hline
Std. Dev. & MI \& STS & Skipgram & gen & 0.689 & 0.831 & -0.142 & 0.055 & 0.443 & -0.388 \\ \hline
Std. Dev. & MI \& STS & all-MiniLM-L12-v2 & clinical & 0.583 & 0.781 & -0.199 & 0.039 & 0.328 & -0.289 \\ \hline
Std. Dev. & MI \& STS & all-MiniLM-L12-v2 & combined & 0.556 & 0.754 & -0.198 & 0.035 & 0.212 & -0.177 \\ \hline
Std. Dev. & MI \& STS & all-MiniLM-L12-v2 & gen & 0.664 & 0.824 & -0.160 & 0.052 & 0.231 & -0.179 \\ \hline
Std. Dev. & MI \& STS & bert-base-uncased & clinical & 0.561 & 0.743 & -0.183 & 0.035 & 0.143 & -0.108 \\ \hline
Std. Dev. & MI \& STS & bert-base-uncased & combined & 0.617 & 0.753 & -0.136 & 0.072 & 0.154 & -0.082 \\ \hline
Std. Dev. & MI \& STS & bert-base-uncased & gen & 0.653 & 0.748 & -0.095 & 0.055 & 0.143 & -0.087 \\ \hline
Std. Dev. & MI \& STS & biomed\_roberta\_base & clinical & 0.719 & 0.711 & 0.008 & 0.088 & 0.149 & -0.061 \\ \hline
Std. Dev. & MI \& STS & biomed\_roberta\_base & combined & 0.680 & 0.769 & -0.089 & 0.047 & 0.187 & -0.139 \\ \hline
Std. Dev. & MI \& STS & biomed\_roberta\_base & gen & 0.705 & 0.746 & -0.041 & 0.052 & 0.127 & -0.075 \\ \hline
Std. Dev. & STS & Bio\_ClinicalBERT & ClinicalSTS & 0.687 & 0.755 & -0.067 & 0.048 & 0.197 & -0.149 \\ \hline
Std. Dev. & STS & Bio\_ClinicalBERT & clinical & 0.668 & 0.727 & -0.059 & 0.046 & 0.218 & -0.172 \\ \hline
Std. Dev. & STS & Bio\_ClinicalBERT & combined & 0.628 & 0.774 & -0.146 & 0.041 & 0.284 & -0.243 \\ \hline
Std. Dev. & STS & Bio\_ClinicalBERT & gen & 0.645 & 0.783 & -0.138 & 0.042 & 0.230 & -0.187 \\ \hline
Std. Dev. & STS & FastText & gen & 0.581 & 0.823 & -0.242 & 0.038 & 0.314 & -0.277 \\ \hline
Std. Dev. & STS & PubMedBERT & clinical & 0.540 & 0.789 & -0.249 & 0.033 & 0.277 & -0.244 \\ \hline
Std. Dev. & STS & PubMedBERT & combined & 0.628 & 0.784 & -0.156 & 0.043 & 0.342 & -0.299 \\ \hline
Std. Dev. & STS & PubMedBERT & gen & 0.647 & 0.790 & -0.143 & 0.045 & 0.326 & -0.281 \\ \hline
Std. Dev. & STS & Skipgram & gen & 0.548 & 0.859 & -0.311 & 0.035 & 0.469 & -0.435 \\ \hline
Std. Dev. & STS & all-MiniLM-L12-v2 & clinical & 0.705 & 0.761 & -0.056 & 0.051 & 0.189 & -0.138 \\ \hline
Std. Dev. & STS & all-MiniLM-L12-v2 & combined & 0.545 & 0.789 & -0.243 & 0.036 & 0.297 & -0.261 \\ \hline
Std. Dev. & STS & all-MiniLM-L12-v2 & gen & 0.678 & 0.798 & -0.120 & 0.048 & 0.274 & -0.226 \\ \hline
Std. Dev. & STS & bert-base-uncased & clinical & 0.590 & 0.774 & -0.184 & 0.037 & 0.272 & -0.235 \\ \hline
Std. Dev. & STS & bert-base-uncased & combined & 0.614 & 0.778 & -0.163 & 0.039 & 0.299 & -0.260 \\ \hline
Std. Dev. & STS & bert-base-uncased & gen & 0.588 & 0.793 & -0.205 & 0.037 & 0.316 & -0.279 \\ \hline
Std. Dev. & STS & biomed\_roberta\_base & clinical & 0.534 & 0.815 & -0.281 & 0.034 & 0.290 & -0.256 \\ \hline
Std. Dev. & STS & biomed\_roberta\_base & combined & 0.347 & \textbf{0.905} & -0.558 & 0.024 & \textbf{0.566} & -0.542 \\ \hline
Std. Dev. & STS & biomed\_roberta\_base & gen & 0.584 & 0.792 & -0.208 & 0.037 & 0.252 & -0.215 \\ \hline
Top N & MI & - & - & 0.331 & 0.521 & -0.190 & 0.023 & 0.048 & -0.025 \\ \hline
Top N & MI \& STS & Bio\_ClinicalBERT & ClinicalSTS & 0.782 & 0.710 & 0.072 & 0.149 & 0.131 & \textbf{0.017} \\ \hline
Top N & MI \& STS & Bio\_ClinicalBERT & clinical & 0.661 & 0.758 & -0.097 & 0.045 & 0.173 & -0.128 \\ \hline
Top N & MI \& STS & Bio\_ClinicalBERT & combined & 0.730 & 0.743 & -0.013 & 0.059 & 0.166 & -0.107 \\ \hline
Top N & MI \& STS & Bio\_ClinicalBERT & gen & 0.675 & 0.772 & -0.097 & 0.049 & 0.188 & -0.140 \\ \hline
Top N & MI \& STS & FastText & gen & 0.171 & 0.504 & -0.333 & 0.019 & 0.046 & -0.027 \\ \hline
Top N & MI \& STS & PubMedBERT & clinical & 0.702 & 0.757 & -0.054 & 0.052 & 0.191 & -0.139 \\ \hline
Top N & MI \& STS & PubMedBERT & combined & \textbf{0.821} & 0.734 & \textbf{0.087} & 0.100 & 0.138 & -0.039 \\ \hline
Top N & MI \& STS & PubMedBERT & gen & 0.727 & 0.744 & -0.016 & 0.058 & 0.209 & -0.150 \\ \hline
Top N & MI \& STS & Skipgram & gen & 0.135 & 0.552 & -0.417 & 0.018 & 0.068 & -0.050 \\ \hline
Top N & MI \& STS & all-MiniLM-L12-v2 & clinical & 0.698 & 0.738 & -0.039 & 0.049 & 0.146 & -0.097 \\ \hline
Top N & MI \& STS & all-MiniLM-L12-v2 & combined & 0.671 & 0.741 & -0.071 & 0.046 & 0.173 & -0.127 \\ \hline
Top N & MI \& STS & all-MiniLM-L12-v2 & gen & 0.769 & 0.722 & 0.047 & 0.065 & 0.137 & -0.071 \\ \hline
Top N & MI \& STS & bert-base-uncased & clinical & 0.654 & 0.715 & -0.061 & 0.047 & 0.089 & -0.041 \\ \hline
Top N & MI \& STS & bert-base-uncased & combined & 0.610 & 0.729 & -0.119 & 0.065 & 0.127 & -0.062 \\ \hline
Top N & MI \& STS & bert-base-uncased & gen & 0.599 & 0.750 & -0.150 & 0.040 & 0.135 & -0.095 \\ \hline
Top N & MI \& STS & biomed\_roberta\_base & clinical & 0.708 & 0.733 & -0.025 & 0.100 & 0.125 & -0.025 \\ \hline
Top N & MI \& STS & biomed\_roberta\_base & combined & 0.760 & 0.729 & 0.032 & 0.062 & 0.150 & -0.088 \\ \hline
Top N & MI \& STS & biomed\_roberta\_base & gen & 0.722 & 0.747 & -0.025 & 0.055 & 0.125 & -0.070 \\ \hline
Top N & STS & Bio\_ClinicalBERT & ClinicalSTS & 0.707 & 0.731 & -0.025 & 0.051 & 0.102 & -0.051 \\ \hline
Top N & STS & Bio\_ClinicalBERT & clinical & 0.702 & 0.739 & -0.037 & 0.049 & 0.154 & -0.105 \\ \hline
Top N & STS & Bio\_ClinicalBERT & combined & 0.669 & 0.730 & -0.061 & 0.045 & 0.162 & -0.117 \\ \hline
Top N & STS & Bio\_ClinicalBERT & gen & 0.664 & 0.738 & -0.074 & 0.045 & 0.156 & -0.111 \\ \hline
Top N & STS & FastText & gen & 0.638 & 0.748 & -0.110 & 0.043 & 0.159 & -0.116 \\ \hline
Top N & STS & PubMedBERT & clinical & 0.736 & 0.738 & -0.003 & 0.055 & 0.168 & -0.112 \\ \hline
Top N & STS & PubMedBERT & combined & 0.736 & 0.737 & \textbf{-0.001} & 0.056 & 0.160 & -0.104 \\ \hline
Top N & STS & PubMedBERT & gen & 0.639 & 0.762 & -0.122 & 0.042 & 0.187 & -0.145 \\ \hline
Top N & STS & Skipgram & gen & 0.730 & 0.762 & -0.032 & 0.057 & 0.168 & -0.110 \\ \hline
Top N & STS & all-MiniLM-L12-v2 & clinical & 0.719 & 0.743 & -0.024 & 0.056 & 0.124 & -0.068 \\ \hline
Top N & STS & all-MiniLM-L12-v2 & combined & 0.595 & 0.749 & -0.154 & 0.038 & 0.167 & -0.129 \\ \hline
Top N & STS & all-MiniLM-L12-v2 & gen & 0.736 & 0.738 & -0.002 & 0.056 & 0.105 & -0.049 \\ \hline
Top N & STS & bert-base-uncased & clinical & 0.640 & 0.723 & -0.082 & 0.046 & 0.160 & -0.114 \\ \hline
Top N & STS & bert-base-uncased & combined & 0.599 & 0.774 & -0.174 & 0.037 & 0.216 & -0.178 \\ \hline
Top N & STS & bert-base-uncased & gen & 0.658 & 0.796 & -0.138 & 0.044 & 0.249 & -0.206 \\ \hline
Top N & STS & biomed\_roberta\_base & clinical & 0.566 & 0.755 & -0.189 & 0.038 & 0.267 & -0.229 \\ \hline
Top N & STS & biomed\_roberta\_base & combined & 0.530 & 0.820 & -0.290 & 0.032 & 0.262 & -0.230 \\ \hline
Top N & STS & biomed\_roberta\_base & gen & 0.653 & 0.744 & -0.091 & 0.044 & 0.121 & -0.077 \\ \hline
mRMR & MI & - & - & 0.430 & 0.591 & -0.161 & 0.031 & 0.075 & -0.044 \\ \hline
mRMR & MI \& STS & Bio\_ClinicalBERT & ClinicalSTS & 0.716 & 0.749 & -0.032 & 0.053 & 0.107 & -0.054 \\ \hline
mRMR & MI \& STS & Bio\_ClinicalBERT & clinical & 0.613 & 0.735 & -0.122 & 0.039 & 0.134 & -0.095 \\ \hline
mRMR & MI \& STS & Bio\_ClinicalBERT & combined & 0.603 & 0.874 & -0.271 & 0.038 & 0.316 & -0.278 \\ \hline
mRMR & MI \& STS & Bio\_ClinicalBERT & gen & 0.671 & 0.740 & -0.069 & 0.046 & 0.157 & -0.111 \\ \hline
mRMR & MI \& STS & FastText & gen & 0.289 & 0.728 & -0.439 & 0.023 & 0.144 & -0.121 \\ \hline
mRMR & MI \& STS & PubMedBERT & clinical & 0.572 & 0.790 & -0.219 & 0.037 & 0.255 & -0.218 \\ \hline
mRMR & MI \& STS & PubMedBERT & combined & 0.632 & 0.825 & -0.193 & 0.052 & 0.387 & -0.336 \\ \hline
mRMR & MI \& STS & PubMedBERT & gen & 0.640 & 0.775 & -0.134 & 0.054 & 0.298 & -0.243 \\ \hline
mRMR & MI \& STS & Skipgram & gen & 0.612 & 0.669 & -0.057 & 0.042 & 0.131 & -0.089 \\ \hline
mRMR & MI \& STS & all-MiniLM-L12-v2 & clinical & 0.609 & 0.728 & -0.119 & 0.040 & 0.121 & -0.081 \\ \hline
mRMR & MI \& STS & all-MiniLM-L12-v2 & combined & 0.606 & 0.850 & -0.244 & 0.039 & 0.226 & -0.187 \\ \hline
mRMR & MI \& STS & all-MiniLM-L12-v2 & gen & 0.763 & 0.727 & 0.036 & 0.071 & 0.115 & -0.044 \\ \hline
mRMR & MI \& STS & bert-base-uncased & clinical & 0.639 & 0.705 & -0.066 & 0.042 & 0.101 & -0.059 \\ \hline
mRMR & MI \& STS & bert-base-uncased & combined & 0.713 & 0.720 & -0.006 & 0.062 & 0.104 & -0.042 \\ \hline
mRMR & MI \& STS & bert-base-uncased & gen & 0.607 & 0.704 & -0.097 & 0.038 & 0.130 & -0.092 \\ \hline
mRMR & MI \& STS & biomed\_roberta\_base & clinical & 0.599 & 0.694 & -0.095 & 0.052 & 0.087 & -0.035 \\ \hline
mRMR & MI \& STS & biomed\_roberta\_base & combined & 0.617 & 0.767 & -0.150 & 0.058 & 0.185 & -0.127 \\ \hline
mRMR & MI \& STS & biomed\_roberta\_base & gen & 0.594 & 0.843 & -0.250 & 0.038 & 0.263 & -0.225 \\ \hline
mRMR & STS & Bio\_ClinicalBERT & ClinicalSTS & 0.713 & 0.738 & -0.024 & 0.074 & 0.189 & -0.115 \\ \hline
mRMR & STS & Bio\_ClinicalBERT & clinical & 0.682 & 0.751 & -0.070 & 0.049 & 0.175 & -0.126 \\ \hline
mRMR & STS & Bio\_ClinicalBERT & combined & 0.529 & 0.760 & -0.231 & 0.033 & 0.236 & -0.203 \\ \hline
mRMR & STS & Bio\_ClinicalBERT & gen & 0.598 & 0.761 & -0.163 & 0.041 & 0.242 & -0.201 \\ \hline
mRMR & STS & FastText & gen & 0.410 & 0.792 & -0.382 & 0.027 & 0.274 & -0.246 \\ \hline
mRMR & STS & PubMedBERT & clinical & 0.587 & 0.835 & -0.248 & 0.038 & 0.277 & -0.239 \\ \hline
mRMR & STS & PubMedBERT & combined & 0.661 & 0.741 & -0.080 & 0.048 & 0.206 & -0.158 \\ \hline
mRMR & STS & PubMedBERT & gen & 0.653 & 0.750 & -0.097 & 0.058 & 0.239 & -0.181 \\ \hline
mRMR & STS & Skipgram & gen & 0.399 & 0.814 & -0.414 & 0.026 & 0.422 & -0.396 \\ \hline
mRMR & STS & all-MiniLM-L12-v2 & clinical & 0.552 & 0.806 & -0.254 & 0.038 & 0.198 & -0.160 \\ \hline
mRMR & STS & all-MiniLM-L12-v2 & combined & 0.563 & 0.745 & -0.182 & 0.035 & 0.145 & -0.110 \\ \hline
mRMR & STS & all-MiniLM-L12-v2 & gen & 0.585 & 0.813 & -0.228 & 0.040 & 0.266 & -0.226 \\ \hline
mRMR & STS & bert-base-uncased & clinical & 0.587 & 0.792 & -0.205 & 0.038 & 0.208 & -0.170 \\ \hline
mRMR & STS & bert-base-uncased & combined & 0.595 & 0.766 & -0.171 & 0.038 & 0.176 & -0.138 \\ \hline
mRMR & STS & bert-base-uncased & gen & 0.603 & 0.746 & -0.142 & 0.040 & 0.110 & -0.070 \\ \hline
mRMR & STS & biomed\_roberta\_base & clinical & 0.532 & 0.818 & -0.286 & 0.035 & 0.223 & -0.188 \\ \hline
mRMR & STS & biomed\_roberta\_base & combined & 0.686 & 0.738 & -0.053 & 0.049 & 0.172 & -0.123 \\ \hline
mRMR & STS & biomed\_roberta\_base & gen & 0.603 & 0.738 & -0.135 & 0.039 & 0.120 & -0.081 \\ \hline
\end{longtable}

\subsection{\gls{gnb}}
\begin{longtable}{|l|l|l|l|l|l|l|l|l|l|}
\caption{All results for \gls{gnb}.} \label{tab:GaussianNB} \\ \hline
\multicolumn{4}{|l|}{Selection Hyperparameters} & \multicolumn{3}{|l|}{AUROC} & \multicolumn{3}{|l|}{AUPRC} \\ \hline
Feature Selector & Scorer & Scoring Model & FT Dataset & Test & Train & $\Delta$ & Test & Train & $\Delta$ \\ \hline
\endfirsthead
\caption[]{All results for \gls{gnb}.} \\ \hline
\multicolumn{4}{|l|}{Selection Hyperparameters} & \multicolumn{3}{|l|}{AUROC} & \multicolumn{3}{|l|}{AUPRC} \\ \hline
Feature Selector & Scorer & Scoring Model & FT Dataset & Test & Train & $\Delta$ & Test & Train & $\Delta$ \\ \hline
\endhead
\hline
\multicolumn{10}{r}{Continued on next page} \\ \hline
\endfoot
\hline
\endlastfoot
Identity & - & - & - & 0.579 & 0.602 & -0.023 & 0.029 & 0.055 & -0.027 \\ \hline
RFE & - & - & - & 0.603 & 0.799 & -0.196 & 0.038 & 0.132 & -0.094 \\ \hline
ReliefF & - & - & - & 0.656 & 0.753 & -0.097 & 0.049 & 0.125 & -0.076 \\ \hline
SFM-LinearSVM & - & - & - & 0.573 & \textbf{0.809} & -0.236 & 0.036 & 0.257 & -0.220 \\ \hline
SFM-XGBoost & - & - & - & 0.559 & 0.761 & -0.202 & 0.036 & 0.236 & -0.200 \\ \hline
Std. Dev. & MI & - & - & 0.533 & 0.565 & -0.032 & 0.026 & 0.051 & -0.025 \\ \hline
Std. Dev. & MI \& STS & Bio\_ClinicalBERT & ClinicalSTS & 0.696 & 0.789 & -0.093 & 0.051 & 0.193 & -0.141 \\ \hline
Std. Dev. & MI \& STS & Bio\_ClinicalBERT & clinical & 0.620 & 0.740 & -0.120 & 0.040 & 0.140 & -0.100 \\ \hline
Std. Dev. & MI \& STS & Bio\_ClinicalBERT & combined & 0.658 & 0.746 & -0.087 & 0.044 & 0.144 & -0.100 \\ \hline
Std. Dev. & MI \& STS & Bio\_ClinicalBERT & gen & 0.639 & 0.754 & -0.115 & 0.042 & 0.181 & -0.139 \\ \hline
Std. Dev. & MI \& STS & FastText & gen & 0.574 & 0.613 & -0.038 & 0.028 & 0.057 & -0.029 \\ \hline
Std. Dev. & MI \& STS & PubMedBERT & clinical & 0.713 & 0.797 & -0.084 & 0.053 & 0.222 & -0.169 \\ \hline
Std. Dev. & MI \& STS & PubMedBERT & combined & 0.713 & 0.781 & -0.068 & 0.055 & 0.201 & -0.146 \\ \hline
Std. Dev. & MI \& STS & PubMedBERT & gen & 0.694 & 0.790 & -0.096 & 0.051 & 0.148 & -0.097 \\ \hline
Std. Dev. & MI \& STS & Skipgram & gen & 0.541 & 0.572 & -0.031 & 0.026 & 0.052 & -0.025 \\ \hline
Std. Dev. & MI \& STS & all-MiniLM-L12-v2 & clinical & 0.693 & 0.786 & -0.093 & 0.051 & 0.204 & -0.153 \\ \hline
Std. Dev. & MI \& STS & all-MiniLM-L12-v2 & combined & 0.669 & 0.753 & -0.084 & 0.052 & 0.159 & -0.108 \\ \hline
Std. Dev. & MI \& STS & all-MiniLM-L12-v2 & gen & 0.713 & 0.746 & -0.032 & 0.081 & 0.109 & -0.028 \\ \hline
Std. Dev. & MI \& STS & bert-base-uncased & clinical & 0.709 & 0.728 & -0.019 & 0.076 & 0.114 & -0.039 \\ \hline
Std. Dev. & MI \& STS & bert-base-uncased & combined & 0.653 & 0.750 & -0.097 & 0.064 & 0.145 & -0.081 \\ \hline
Std. Dev. & MI \& STS & bert-base-uncased & gen & 0.700 & 0.742 & -0.042 & 0.082 & 0.115 & -0.033 \\ \hline
Std. Dev. & MI \& STS & biomed\_roberta\_base & clinical & 0.669 & 0.703 & -0.033 & 0.059 & 0.127 & -0.067 \\ \hline
Std. Dev. & MI \& STS & biomed\_roberta\_base & combined & 0.678 & 0.746 & -0.068 & 0.051 & 0.147 & -0.096 \\ \hline
Std. Dev. & MI \& STS & biomed\_roberta\_base & gen & 0.689 & 0.742 & -0.054 & 0.047 & 0.114 & -0.067 \\ \hline
Std. Dev. & STS & Bio\_ClinicalBERT & ClinicalSTS & 0.711 & 0.742 & -0.031 & 0.054 & 0.150 & -0.096 \\ \hline
Std. Dev. & STS & Bio\_ClinicalBERT & clinical & 0.672 & 0.737 & -0.065 & 0.044 & 0.146 & -0.102 \\ \hline
Std. Dev. & STS & Bio\_ClinicalBERT & combined & 0.697 & 0.752 & -0.055 & 0.053 & 0.168 & -0.115 \\ \hline
Std. Dev. & STS & Bio\_ClinicalBERT & gen & 0.680 & 0.745 & -0.065 & 0.051 & 0.153 & -0.103 \\ \hline
Std. Dev. & STS & FastText & gen & 0.731 & 0.708 & 0.024 & 0.044 & 0.075 & -0.031 \\ \hline
Std. Dev. & STS & PubMedBERT & clinical & 0.653 & 0.770 & -0.117 & 0.045 & 0.198 & -0.154 \\ \hline
Std. Dev. & STS & PubMedBERT & combined & 0.661 & 0.778 & -0.117 & 0.044 & 0.202 & -0.157 \\ \hline
Std. Dev. & STS & PubMedBERT & gen & 0.696 & 0.783 & -0.088 & 0.051 & 0.147 & -0.096 \\ \hline
Std. Dev. & STS & Skipgram & gen & \textbf{0.756} & 0.713 & 0.043 & 0.048 & 0.075 & -0.027 \\ \hline
Std. Dev. & STS & all-MiniLM-L12-v2 & clinical & 0.704 & 0.751 & -0.047 & 0.058 & 0.136 & -0.078 \\ \hline
Std. Dev. & STS & all-MiniLM-L12-v2 & combined & 0.687 & 0.760 & -0.073 & 0.050 & 0.173 & -0.123 \\ \hline
Std. Dev. & STS & all-MiniLM-L12-v2 & gen & 0.696 & 0.762 & -0.066 & 0.054 & 0.145 & -0.091 \\ \hline
Std. Dev. & STS & bert-base-uncased & clinical & 0.664 & 0.754 & -0.090 & 0.046 & 0.199 & -0.154 \\ \hline
Std. Dev. & STS & bert-base-uncased & combined & 0.722 & 0.774 & -0.052 & 0.055 & 0.139 & -0.084 \\ \hline
Std. Dev. & STS & bert-base-uncased & gen & 0.679 & 0.780 & -0.101 & 0.048 & 0.141 & -0.093 \\ \hline
Std. Dev. & STS & biomed\_roberta\_base & clinical & 0.726 & 0.775 & -0.049 & 0.048 & 0.113 & -0.065 \\ \hline
Std. Dev. & STS & biomed\_roberta\_base & combined & 0.689 & 0.783 & -0.094 & 0.051 & 0.227 & -0.176 \\ \hline
Std. Dev. & STS & biomed\_roberta\_base & gen & 0.656 & 0.778 & -0.123 & 0.044 & 0.245 & -0.200 \\ \hline
Top N & MI & - & - & 0.521 & 0.535 & -0.014 & 0.025 & 0.048 & -0.023 \\ \hline
Top N & MI \& STS & Bio\_ClinicalBERT & ClinicalSTS & 0.708 & 0.709 & \textbf{-0.001} & \textbf{0.195} & 0.108 & \textbf{0.087} \\ \hline
Top N & MI \& STS & Bio\_ClinicalBERT & clinical & 0.628 & 0.742 & -0.114 & 0.041 & 0.171 & -0.130 \\ \hline
Top N & MI \& STS & Bio\_ClinicalBERT & combined & 0.691 & 0.718 & -0.027 & 0.049 & 0.146 & -0.096 \\ \hline
Top N & MI \& STS & Bio\_ClinicalBERT & gen & 0.628 & 0.737 & -0.109 & 0.041 & 0.154 & -0.113 \\ \hline
Top N & MI \& STS & FastText & gen & 0.493 & 0.739 & -0.246 & 0.035 & 0.132 & -0.097 \\ \hline
Top N & MI \& STS & PubMedBERT & clinical & 0.628 & 0.759 & -0.131 & 0.041 & 0.163 & -0.122 \\ \hline
Top N & MI \& STS & PubMedBERT & combined & 0.738 & 0.767 & -0.028 & 0.075 & 0.145 & -0.070 \\ \hline
Top N & MI \& STS & PubMedBERT & gen & 0.671 & 0.776 & -0.105 & 0.052 & 0.195 & -0.143 \\ \hline
Top N & MI \& STS & Skipgram & gen & 0.515 & 0.750 & -0.235 & 0.038 & 0.134 & -0.096 \\ \hline
Top N & MI \& STS & all-MiniLM-L12-v2 & clinical & 0.629 & 0.743 & -0.113 & 0.041 & 0.144 & -0.103 \\ \hline
Top N & MI \& STS & all-MiniLM-L12-v2 & combined & 0.702 & 0.736 & -0.034 & 0.056 & 0.128 & -0.072 \\ \hline
Top N & MI \& STS & all-MiniLM-L12-v2 & gen & 0.727 & 0.747 & -0.019 & 0.090 & 0.105 & -0.015 \\ \hline
Top N & MI \& STS & bert-base-uncased & clinical & 0.718 & 0.697 & 0.021 & 0.098 & 0.090 & 0.008 \\ \hline
Top N & MI \& STS & bert-base-uncased & combined & 0.616 & 0.734 & -0.119 & 0.050 & 0.123 & -0.073 \\ \hline
Top N & MI \& STS & bert-base-uncased & gen & 0.690 & 0.737 & -0.047 & 0.058 & 0.119 & -0.061 \\ \hline
Top N & MI \& STS & biomed\_roberta\_base & clinical & 0.675 & 0.698 & -0.023 & 0.056 & 0.136 & -0.080 \\ \hline
Top N & MI \& STS & biomed\_roberta\_base & combined & 0.675 & 0.752 & -0.077 & 0.046 & 0.140 & -0.093 \\ \hline
Top N & MI \& STS & biomed\_roberta\_base & gen & 0.689 & 0.742 & -0.054 & 0.047 & 0.114 & -0.067 \\ \hline
Top N & STS & Bio\_ClinicalBERT & ClinicalSTS & 0.647 & 0.741 & -0.094 & 0.042 & 0.105 & -0.063 \\ \hline
Top N & STS & Bio\_ClinicalBERT & clinical & 0.628 & 0.739 & -0.111 & 0.040 & 0.142 & -0.102 \\ \hline
Top N & STS & Bio\_ClinicalBERT & combined & 0.617 & 0.728 & -0.111 & 0.039 & 0.131 & -0.091 \\ \hline
Top N & STS & Bio\_ClinicalBERT & gen & 0.618 & 0.733 & -0.115 & 0.040 & 0.132 & -0.092 \\ \hline
Top N & STS & FastText & gen & 0.617 & 0.744 & -0.127 & 0.040 & 0.146 & -0.105 \\ \hline
Top N & STS & PubMedBERT & clinical & 0.664 & 0.732 & -0.068 & 0.046 & 0.129 & -0.083 \\ \hline
Top N & STS & PubMedBERT & combined & 0.678 & 0.757 & -0.080 & 0.048 & 0.138 & -0.091 \\ \hline
Top N & STS & PubMedBERT & gen & 0.707 & 0.776 & -0.070 & 0.054 & 0.189 & -0.135 \\ \hline
Top N & STS & Skipgram & gen & 0.639 & 0.749 & -0.110 & 0.043 & 0.142 & -0.099 \\ \hline
Top N & STS & all-MiniLM-L12-v2 & clinical & 0.658 & 0.746 & -0.087 & 0.044 & 0.110 & -0.067 \\ \hline
Top N & STS & all-MiniLM-L12-v2 & combined & 0.680 & 0.746 & -0.065 & 0.048 & 0.129 & -0.081 \\ \hline
Top N & STS & all-MiniLM-L12-v2 & gen & 0.653 & 0.738 & -0.085 & 0.043 & 0.111 & -0.068 \\ \hline
Top N & STS & bert-base-uncased & clinical & 0.618 & 0.744 & -0.125 & 0.040 & 0.147 & -0.106 \\ \hline
Top N & STS & bert-base-uncased & combined & 0.687 & 0.743 & -0.055 & 0.046 & 0.170 & -0.124 \\ \hline
Top N & STS & bert-base-uncased & gen & 0.675 & 0.737 & -0.062 & 0.048 & 0.145 & -0.096 \\ \hline
Top N & STS & biomed\_roberta\_base & clinical & 0.613 & 0.727 & -0.114 & 0.043 & 0.171 & -0.128 \\ \hline
Top N & STS & biomed\_roberta\_base & combined & 0.731 & 0.737 & -0.006 & 0.065 & 0.132 & -0.067 \\ \hline
Top N & STS & biomed\_roberta\_base & gen & 0.676 & 0.737 & -0.060 & 0.048 & 0.139 & -0.091 \\ \hline
mRMR & MI & - & - & 0.375 & 0.712 & -0.337 & 0.025 & 0.122 & -0.097 \\ \hline
mRMR & MI \& STS & Bio\_ClinicalBERT & ClinicalSTS & 0.643 & 0.741 & -0.097 & 0.042 & 0.116 & -0.074 \\ \hline
mRMR & MI \& STS & Bio\_ClinicalBERT & clinical & 0.602 & 0.721 & -0.119 & 0.037 & 0.126 & -0.089 \\ \hline
mRMR & MI \& STS & Bio\_ClinicalBERT & combined & 0.573 & 0.732 & -0.159 & 0.036 & 0.134 & -0.099 \\ \hline
mRMR & MI \& STS & Bio\_ClinicalBERT & gen & 0.616 & 0.741 & -0.125 & 0.039 & 0.113 & -0.074 \\ \hline
mRMR & MI \& STS & FastText & gen & 0.482 & 0.700 & -0.218 & 0.031 & 0.197 & -0.166 \\ \hline
mRMR & MI \& STS & PubMedBERT & clinical & 0.680 & 0.727 & -0.047 & 0.056 & 0.143 & -0.087 \\ \hline
mRMR & MI \& STS & PubMedBERT & combined & 0.674 & 0.756 & -0.083 & 0.059 & 0.146 & -0.087 \\ \hline
mRMR & MI \& STS & PubMedBERT & gen & 0.652 & 0.796 & -0.144 & 0.043 & 0.271 & -0.228 \\ \hline
mRMR & MI \& STS & Skipgram & gen & 0.738 & 0.714 & 0.025 & 0.058 & 0.111 & -0.053 \\ \hline
mRMR & MI \& STS & all-MiniLM-L12-v2 & clinical & 0.656 & 0.735 & -0.080 & 0.048 & 0.102 & -0.054 \\ \hline
mRMR & MI \& STS & all-MiniLM-L12-v2 & combined & 0.675 & 0.751 & -0.076 & 0.047 & 0.112 & -0.066 \\ \hline
mRMR & MI \& STS & all-MiniLM-L12-v2 & gen & 0.715 & 0.741 & -0.026 & 0.055 & 0.134 & -0.079 \\ \hline
mRMR & MI \& STS & bert-base-uncased & clinical & 0.700 & 0.701 & \textbf{-0.001} & 0.054 & 0.125 & -0.071 \\ \hline
mRMR & MI \& STS & bert-base-uncased & combined & 0.738 & 0.713 & 0.025 & 0.067 & 0.110 & -0.042 \\ \hline
mRMR & MI \& STS & bert-base-uncased & gen & 0.632 & 0.702 & -0.069 & 0.043 & 0.122 & -0.079 \\ \hline
mRMR & MI \& STS & biomed\_roberta\_base & clinical & 0.638 & 0.712 & -0.074 & 0.045 & 0.085 & -0.039 \\ \hline
mRMR & MI \& STS & biomed\_roberta\_base & combined & 0.749 & 0.698 & \textbf{0.052} & 0.092 & 0.095 & \textbf{-0.004} \\ \hline
mRMR & MI \& STS & biomed\_roberta\_base & gen & 0.671 & 0.751 & -0.080 & 0.046 & 0.108 & -0.062 \\ \hline
mRMR & STS & Bio\_ClinicalBERT & ClinicalSTS & 0.716 & 0.734 & -0.018 & 0.088 & 0.116 & -0.028 \\ \hline
mRMR & STS & Bio\_ClinicalBERT & clinical & 0.652 & 0.755 & -0.104 & 0.043 & 0.125 & -0.082 \\ \hline
mRMR & STS & Bio\_ClinicalBERT & combined & 0.628 & 0.735 & -0.107 & 0.042 & 0.177 & -0.134 \\ \hline
mRMR & STS & Bio\_ClinicalBERT & gen & 0.599 & 0.752 & -0.153 & 0.038 & 0.164 & -0.126 \\ \hline
mRMR & STS & FastText & gen & 0.518 & 0.755 & -0.237 & 0.035 & 0.227 & -0.193 \\ \hline
mRMR & STS & PubMedBERT & clinical & 0.627 & 0.757 & -0.131 & 0.041 & 0.174 & -0.133 \\ \hline
mRMR & STS & PubMedBERT & combined & 0.617 & 0.735 & -0.118 & 0.040 & 0.159 & -0.119 \\ \hline
mRMR & STS & PubMedBERT & gen & 0.656 & 0.743 & -0.087 & 0.047 & 0.224 & -0.177 \\ \hline
mRMR & STS & Skipgram & gen & 0.521 & 0.801 & -0.280 & 0.034 & \textbf{0.327} & -0.293 \\ \hline
mRMR & STS & all-MiniLM-L12-v2 & clinical & 0.663 & 0.747 & -0.084 & 0.048 & 0.113 & -0.065 \\ \hline
mRMR & STS & all-MiniLM-L12-v2 & combined & 0.702 & 0.743 & -0.041 & 0.068 & 0.115 & -0.047 \\ \hline
mRMR & STS & all-MiniLM-L12-v2 & gen & 0.698 & 0.749 & -0.050 & 0.072 & 0.113 & -0.041 \\ \hline
mRMR & STS & bert-base-uncased & clinical & 0.658 & 0.756 & -0.097 & 0.045 & 0.121 & -0.077 \\ \hline
mRMR & STS & bert-base-uncased & combined & 0.647 & 0.776 & -0.129 & 0.043 & 0.157 & -0.114 \\ \hline
mRMR & STS & bert-base-uncased & gen & 0.658 & 0.739 & -0.081 & 0.045 & 0.110 & -0.065 \\ \hline
mRMR & STS & biomed\_roberta\_base & clinical & 0.620 & 0.770 & -0.150 & 0.043 & 0.134 & -0.091 \\ \hline
mRMR & STS & biomed\_roberta\_base & combined & 0.719 & 0.730 & -0.011 & 0.115 & 0.107 & 0.008 \\ \hline
mRMR & STS & biomed\_roberta\_base & gen & 0.683 & 0.722 & -0.039 & 0.066 & 0.110 & -0.044 \\ \hline
\end{longtable}

\subsection{\gls{knn}}
\begin{longtable}{|l|l|l|l|l|l|l|l|l|l|}
\caption{All results for \gls{knn}.} \label{tab:KNearest} \\ \hline
\multicolumn{4}{|l|}{Selection Hyperparameters} & \multicolumn{3}{|l|}{AUROC} & \multicolumn{3}{|l|}{AUPRC} \\ \hline
Feature Selector & Scorer & Scoring Model & FT Dataset & Test & Train & $\Delta$ & Test & Train & $\Delta$ \\ \hline
\endfirsthead
\caption[]{All results for KNearest.} \\ \hline
\multicolumn{4}{|l|}{Selection Hyperparameters} & \multicolumn{3}{|l|}{AUROC} & \multicolumn{3}{|l|}{AUPRC} \\ \hline
Feature Selector & Scorer & Scoring Model & FT Dataset & Test & Train & $\Delta$ & Test & Train & $\Delta$ \\ \hline
\endhead
\hline
\multicolumn{10}{r}{Continued on next page} \\ \hline
\endfoot
\hline
\endlastfoot
Identity & - & - & - & 0.393 & 0.900 & -0.508 & 0.024 & 0.207 & -0.182 \\ \hline
RFE & - & - & - & 0.552 & 0.948 & -0.396 & 0.029 & 0.408 & -0.379 \\ \hline
ReliefF & - & - & - & 0.405 & 0.933 & -0.528 & 0.024 & 0.270 & -0.245 \\ \hline
SFM-LinearSVM & - & - & - & 0.450 & \textbf{0.967} & -0.517 & 0.024 & 0.435 & -0.411 \\ \hline
SFM-XGBoost & - & - & - & 0.588 & 0.908 & -0.320 & 0.099 & 0.210 & -0.110 \\ \hline
Std. Dev. & MI & - & - & 0.744 & 0.908 & -0.164 & 0.072 & 0.214 & -0.143 \\ \hline
Std. Dev. & MI \& STS & Bio\_ClinicalBERT & ClinicalSTS & 0.405 & 0.922 & -0.517 & 0.024 & 0.246 & -0.222 \\ \hline
Std. Dev. & MI \& STS & Bio\_ClinicalBERT & clinical & 0.351 & 0.908 & -0.557 & 0.024 & 0.231 & -0.207 \\ \hline
Std. Dev. & MI \& STS & Bio\_ClinicalBERT & combined & 0.515 & 0.881 & -0.366 & 0.026 & 0.165 & -0.139 \\ \hline
Std. Dev. & MI \& STS & Bio\_ClinicalBERT & gen & 0.567 & 0.919 & -0.352 & 0.030 & 0.230 & -0.200 \\ \hline
Std. Dev. & MI \& STS & FastText & gen & 0.446 & 0.954 & -0.508 & 0.024 & 0.339 & -0.315 \\ \hline
Std. Dev. & MI \& STS & PubMedBERT & clinical & 0.552 & 0.909 & -0.356 & 0.028 & 0.259 & -0.231 \\ \hline
Std. Dev. & MI \& STS & PubMedBERT & combined & 0.430 & 0.953 & -0.524 & 0.024 & 0.333 & -0.309 \\ \hline
Std. Dev. & MI \& STS & PubMedBERT & gen & 0.529 & 0.912 & -0.383 & 0.026 & 0.349 & -0.322 \\ \hline
Std. Dev. & MI \& STS & Skipgram & gen & 0.585 & 0.927 & -0.342 & 0.099 & 0.316 & -0.216 \\ \hline
Std. Dev. & MI \& STS & all-MiniLM-L12-v2 & clinical & 0.459 & 0.923 & -0.464 & 0.024 & 0.404 & -0.380 \\ \hline
Std. Dev. & MI \& STS & all-MiniLM-L12-v2 & combined & 0.446 & 0.956 & -0.510 & 0.024 & 0.364 & -0.339 \\ \hline
Std. Dev. & MI \& STS & all-MiniLM-L12-v2 & gen & 0.689 & 0.896 & -0.207 & 0.051 & 0.184 & -0.133 \\ \hline
Std. Dev. & MI \& STS & bert-base-uncased & clinical & 0.562 & 0.879 & -0.317 & 0.029 & 0.208 & -0.179 \\ \hline
Std. Dev. & MI \& STS & bert-base-uncased & combined & 0.426 & 0.933 & -0.507 & 0.024 & 0.322 & -0.298 \\ \hline
Std. Dev. & MI \& STS & bert-base-uncased & gen & 0.696 & 0.904 & -0.209 & 0.050 & 0.251 & -0.201 \\ \hline
Std. Dev. & MI \& STS & biomed\_roberta\_base & clinical & 0.588 & 0.958 & -0.369 & 0.033 & 0.394 & -0.361 \\ \hline
Std. Dev. & MI \& STS & biomed\_roberta\_base & combined & 0.580 & 0.958 & -0.379 & 0.032 & 0.373 & -0.341 \\ \hline
Std. Dev. & MI \& STS & biomed\_roberta\_base & gen & 0.696 & 0.799 & -0.103 & 0.047 & 0.189 & -0.142 \\ \hline
Std. Dev. & STS & Bio\_ClinicalBERT & ClinicalSTS & 0.446 & 0.820 & -0.373 & 0.024 & 0.252 & -0.228 \\ \hline
Std. Dev. & STS & Bio\_ClinicalBERT & clinical & 0.758 & 0.777 & \textbf{-0.019} & 0.078 & 0.225 & -0.147 \\ \hline
Std. Dev. & STS & Bio\_ClinicalBERT & combined & 0.364 & 0.897 & -0.533 & 0.024 & 0.211 & -0.186 \\ \hline
Std. Dev. & STS & Bio\_ClinicalBERT & gen & 0.438 & 0.928 & -0.490 & 0.024 & 0.312 & -0.288 \\ \hline
Std. Dev. & STS & FastText & gen & 0.561 & 0.908 & -0.347 & 0.029 & 0.226 & -0.197 \\ \hline
Std. Dev. & STS & PubMedBERT & clinical & 0.588 & 0.948 & -0.360 & 0.034 & 0.428 & -0.394 \\ \hline
Std. Dev. & STS & PubMedBERT & combined & 0.397 & 0.928 & -0.531 & 0.024 & 0.296 & -0.271 \\ \hline
Std. Dev. & STS & PubMedBERT & gen & 0.548 & 0.905 & -0.357 & 0.028 & 0.238 & -0.210 \\ \hline
Std. Dev. & STS & Skipgram & gen & 0.417 & 0.937 & -0.520 & 0.024 & 0.298 & -0.274 \\ \hline
Std. Dev. & STS & all-MiniLM-L12-v2 & clinical & 0.672 & 0.864 & -0.192 & 0.044 & 0.177 & -0.133 \\ \hline
Std. Dev. & STS & all-MiniLM-L12-v2 & combined & 0.393 & 0.910 & -0.518 & 0.024 & 0.298 & -0.274 \\ \hline
Std. Dev. & STS & all-MiniLM-L12-v2 & gen & 0.584 & 0.921 & -0.337 & 0.033 & 0.243 & -0.210 \\ \hline
Std. Dev. & STS & bert-base-uncased & clinical & 0.413 & 0.890 & -0.477 & 0.024 & 0.234 & -0.210 \\ \hline
Std. Dev. & STS & bert-base-uncased & combined & 0.434 & 0.928 & -0.494 & 0.024 & 0.359 & -0.335 \\ \hline
Std. Dev. & STS & bert-base-uncased & gen & 0.442 & 0.934 & -0.492 & 0.024 & 0.412 & -0.388 \\ \hline
Std. Dev. & STS & biomed\_roberta\_base & clinical & 0.555 & 0.936 & -0.381 & 0.028 & 0.312 & -0.284 \\ \hline
Std. Dev. & STS & biomed\_roberta\_base & combined & 0.368 & 0.909 & -0.541 & 0.024 & 0.225 & -0.201 \\ \hline
Std. Dev. & STS & biomed\_roberta\_base & gen & 0.360 & 0.884 & -0.525 & 0.024 & 0.178 & -0.153 \\ \hline
Top N & MI & - & - & 0.569 & 0.930 & -0.361 & 0.030 & 0.253 & -0.223 \\ \hline
Top N & MI \& STS & Bio\_ClinicalBERT & ClinicalSTS & \textbf{0.784} & 0.724 & \textbf{0.060} & \textbf{0.103} & 0.148 & \textbf{-0.045} \\ \hline
Top N & MI \& STS & Bio\_ClinicalBERT & clinical & 0.376 & 0.912 & -0.536 & 0.024 & 0.236 & -0.212 \\ \hline
Top N & MI \& STS & Bio\_ClinicalBERT & combined & 0.372 & 0.903 & -0.531 & 0.024 & 0.213 & -0.189 \\ \hline
Top N & MI \& STS & Bio\_ClinicalBERT & gen & 0.409 & 0.939 & -0.529 & 0.024 & 0.300 & -0.276 \\ \hline
Top N & MI \& STS & FastText & gen & 0.629 & 0.954 & -0.325 & 0.049 & 0.400 & -0.351 \\ \hline
Top N & MI \& STS & PubMedBERT & clinical & 0.421 & 0.941 & -0.520 & 0.024 & 0.396 & -0.371 \\ \hline
Top N & MI \& STS & PubMedBERT & combined & 0.769 & 0.942 & -0.173 & 0.086 & 0.406 & -0.319 \\ \hline
Top N & MI \& STS & PubMedBERT & gen & 0.351 & 0.865 & -0.514 & 0.024 & 0.207 & -0.183 \\ \hline
Top N & MI \& STS & Skipgram & gen & 0.592 & 0.950 & -0.357 & 0.034 & 0.383 & -0.349 \\ \hline
Top N & MI \& STS & all-MiniLM-L12-v2 & clinical & 0.716 & 0.807 & -0.091 & 0.054 & 0.154 & -0.100 \\ \hline
Top N & MI \& STS & all-MiniLM-L12-v2 & combined & 0.558 & 0.788 & -0.231 & 0.029 & 0.151 & -0.122 \\ \hline
Top N & MI \& STS & all-MiniLM-L12-v2 & gen & 0.413 & 0.930 & -0.516 & 0.024 & 0.310 & -0.286 \\ \hline
Top N & MI \& STS & bert-base-uncased & clinical & 0.723 & 0.819 & -0.096 & 0.056 & 0.178 & -0.122 \\ \hline
Top N & MI \& STS & bert-base-uncased & combined & 0.607 & 0.808 & -0.200 & 0.038 & 0.311 & -0.273 \\ \hline
Top N & MI \& STS & bert-base-uncased & gen & 0.599 & 0.821 & -0.222 & 0.036 & 0.245 & -0.210 \\ \hline
Top N & MI \& STS & biomed\_roberta\_base & clinical & 0.596 & 0.955 & -0.358 & 0.035 & 0.355 & -0.320 \\ \hline
Top N & MI \& STS & biomed\_roberta\_base & combined & 0.426 & 0.934 & -0.508 & 0.024 & 0.328 & -0.304 \\ \hline
Top N & MI \& STS & biomed\_roberta\_base & gen & 0.696 & 0.799 & -0.103 & 0.047 & 0.189 & -0.142 \\ \hline
Top N & STS & Bio\_ClinicalBERT & ClinicalSTS & 0.550 & 0.753 & -0.204 & 0.028 & 0.189 & -0.161 \\ \hline
Top N & STS & Bio\_ClinicalBERT & clinical & 0.691 & 0.785 & -0.094 & 0.046 & 0.152 & -0.106 \\ \hline
Top N & STS & Bio\_ClinicalBERT & combined & 0.601 & 0.715 & -0.114 & 0.037 & 0.179 & -0.142 \\ \hline
Top N & STS & Bio\_ClinicalBERT & gen & 0.540 & 0.717 & -0.177 & 0.027 & 0.119 & -0.092 \\ \hline
Top N & STS & FastText & gen & 0.417 & 0.879 & -0.461 & 0.024 & 0.289 & -0.265 \\ \hline
Top N & STS & PubMedBERT & clinical & 0.562 & 0.805 & -0.243 & 0.040 & 0.139 & -0.099 \\ \hline
Top N & STS & PubMedBERT & combined & 0.558 & 0.810 & -0.252 & 0.040 & 0.141 & -0.101 \\ \hline
Top N & STS & PubMedBERT & gen & 0.376 & 0.871 & -0.495 & 0.024 & 0.260 & -0.236 \\ \hline
Top N & STS & Skipgram & gen & 0.405 & 0.894 & -0.489 & 0.024 & 0.238 & -0.214 \\ \hline
Top N & STS & all-MiniLM-L12-v2 & clinical & 0.674 & 0.778 & -0.105 & 0.043 & 0.160 & -0.117 \\ \hline
Top N & STS & all-MiniLM-L12-v2 & combined & 0.360 & 0.888 & -0.528 & 0.024 & 0.211 & -0.187 \\ \hline
Top N & STS & all-MiniLM-L12-v2 & gen & 0.566 & 0.752 & -0.186 & 0.029 & 0.201 & -0.171 \\ \hline
Top N & STS & bert-base-uncased & clinical & 0.585 & 0.808 & -0.223 & 0.033 & 0.259 & -0.227 \\ \hline
Top N & STS & bert-base-uncased & combined & 0.405 & 0.780 & -0.375 & 0.024 & 0.223 & -0.199 \\ \hline
Top N & STS & bert-base-uncased & gen & 0.503 & 0.840 & -0.337 & 0.025 & 0.202 & -0.177 \\ \hline
Top N & STS & biomed\_roberta\_base & clinical & 0.372 & 0.890 & -0.518 & 0.024 & 0.264 & -0.240 \\ \hline
Top N & STS & biomed\_roberta\_base & combined & 0.355 & 0.897 & -0.542 & 0.024 & 0.220 & -0.196 \\ \hline
Top N & STS & biomed\_roberta\_base & gen & 0.624 & 0.667 & -0.043 & 0.046 & 0.143 & -0.097 \\ \hline
mRMR & MI & - & - & 0.397 & 0.924 & -0.527 & 0.024 & 0.279 & -0.255 \\ \hline
mRMR & MI \& STS & Bio\_ClinicalBERT & ClinicalSTS & 0.541 & 0.802 & -0.260 & 0.027 & 0.198 & -0.171 \\ \hline
mRMR & MI \& STS & Bio\_ClinicalBERT & clinical & 0.434 & 0.837 & -0.403 & 0.024 & 0.243 & -0.219 \\ \hline
mRMR & MI \& STS & Bio\_ClinicalBERT & combined & 0.364 & 0.849 & -0.485 & 0.024 & 0.184 & -0.160 \\ \hline
mRMR & MI \& STS & Bio\_ClinicalBERT & gen & 0.426 & 0.862 & -0.437 & 0.024 & 0.245 & -0.221 \\ \hline
mRMR & MI \& STS & FastText & gen & 0.401 & 0.932 & -0.532 & 0.024 & 0.287 & -0.263 \\ \hline
mRMR & MI \& STS & PubMedBERT & clinical & 0.573 & 0.826 & -0.253 & 0.083 & 0.149 & -0.067 \\ \hline
mRMR & MI \& STS & PubMedBERT & combined & 0.687 & 0.820 & -0.132 & 0.050 & 0.228 & -0.178 \\ \hline
mRMR & MI \& STS & PubMedBERT & gen & 0.529 & 0.882 & -0.353 & 0.027 & 0.306 & -0.280 \\ \hline
mRMR & MI \& STS & Skipgram & gen & 0.551 & 0.919 & -0.368 & 0.028 & 0.238 & -0.210 \\ \hline
mRMR & MI \& STS & all-MiniLM-L12-v2 & clinical & 0.518 & 0.786 & -0.268 & 0.025 & 0.119 & -0.094 \\ \hline
mRMR & MI \& STS & all-MiniLM-L12-v2 & combined & 0.570 & 0.853 & -0.283 & 0.030 & 0.262 & -0.232 \\ \hline
mRMR & MI \& STS & all-MiniLM-L12-v2 & gen & 0.645 & 0.786 & -0.141 & 0.039 & 0.162 & -0.123 \\ \hline
mRMR & MI \& STS & bert-base-uncased & clinical & 0.624 & 0.768 & -0.144 & 0.046 & 0.191 & -0.145 \\ \hline
mRMR & MI \& STS & bert-base-uncased & combined & 0.463 & 0.766 & -0.303 & 0.024 & 0.206 & -0.181 \\ \hline
mRMR & MI \& STS & bert-base-uncased & gen & 0.388 & 0.790 & -0.402 & 0.024 & 0.157 & -0.132 \\ \hline
mRMR & MI \& STS & biomed\_roberta\_base & clinical & 0.475 & 0.748 & -0.272 & 0.024 & 0.151 & -0.127 \\ \hline
mRMR & MI \& STS & biomed\_roberta\_base & combined & 0.515 & 0.895 & -0.380 & 0.025 & 0.213 & -0.188 \\ \hline
mRMR & MI \& STS & biomed\_roberta\_base & gen & 0.559 & 0.874 & -0.315 & 0.029 & 0.270 & -0.241 \\ \hline
mRMR & STS & Bio\_ClinicalBERT & ClinicalSTS & 0.526 & 0.898 & -0.372 & 0.026 & 0.237 & -0.211 \\ \hline
mRMR & STS & Bio\_ClinicalBERT & clinical & 0.397 & 0.850 & -0.453 & 0.024 & 0.180 & -0.156 \\ \hline
mRMR & STS & Bio\_ClinicalBERT & combined & 0.397 & 0.915 & -0.518 & 0.024 & 0.253 & -0.229 \\ \hline
mRMR & STS & Bio\_ClinicalBERT & gen & 0.393 & 0.873 & -0.481 & 0.024 & 0.219 & -0.195 \\ \hline
mRMR & STS & FastText & gen & 0.455 & \textbf{0.967} & -0.512 & 0.024 & \textbf{0.436} & -0.411 \\ \hline
mRMR & STS & PubMedBERT & clinical & 0.569 & 0.901 & -0.332 & 0.030 & 0.328 & -0.298 \\ \hline
mRMR & STS & PubMedBERT & combined & 0.372 & 0.907 & -0.535 & 0.024 & 0.215 & -0.191 \\ \hline
mRMR & STS & PubMedBERT & gen & 0.368 & 0.902 & -0.534 & 0.024 & 0.212 & -0.188 \\ \hline
mRMR & STS & Skipgram & gen & 0.388 & 0.928 & -0.539 & 0.024 & 0.302 & -0.278 \\ \hline
mRMR & STS & all-MiniLM-L12-v2 & clinical & 0.584 & 0.884 & -0.300 & 0.083 & 0.212 & -0.129 \\ \hline
mRMR & STS & all-MiniLM-L12-v2 & combined & 0.506 & 0.873 & -0.368 & 0.025 & 0.218 & -0.193 \\ \hline
mRMR & STS & all-MiniLM-L12-v2 & gen & 0.426 & 0.882 & -0.457 & 0.024 & 0.226 & -0.202 \\ \hline
mRMR & STS & bert-base-uncased & clinical & 0.413 & 0.940 & -0.527 & 0.024 & 0.299 & -0.275 \\ \hline
mRMR & STS & bert-base-uncased & combined & 0.401 & 0.926 & -0.525 & 0.024 & 0.274 & -0.250 \\ \hline
mRMR & STS & bert-base-uncased & gen & 0.536 & 0.891 & -0.355 & 0.027 & 0.188 & -0.161 \\ \hline
mRMR & STS & biomed\_roberta\_base & clinical & 0.442 & 0.946 & -0.504 & 0.024 & 0.324 & -0.299 \\ \hline
mRMR & STS & biomed\_roberta\_base & combined & 0.409 & 0.932 & -0.523 & 0.024 & 0.261 & -0.237 \\ \hline
mRMR & STS & biomed\_roberta\_base & gen & 0.446 & 0.778 & -0.332 & 0.024 & 0.193 & -0.169 \\ \hline
\end{longtable}